%% file: sample-sigconf.tex
\documentclass[sigconf,natbib=true,screen]{acmart}
\usepackage{xspace,balance,tabularx,multirow}
\usepackage{flushend}
\usepackage{tikz}
\usepackage{pgfplots}
\pgfplotsset{compat=1.16}
\usetikzlibrary{patterns}
\usepackage{subfig}
\usepackage{amsmath,amsfonts}
\usepackage{algorithm}
\usepackage{algpseudocode}
\usepackage{xcolor}
\usepackage{colortbl}
\usepackage{bbold}
\usepackage{enumitem}
\usepackage{tablefootnote}
\usepackage{upgreek,textgreek}
\usepackage{pifont}
\usepackage[noabbrev]{cleveref}
\usepackage{titlecaps}
\usepackage{lipsum}
\usepackage{makecell}
\usepackage{stfloats}

\pgfplotsset{every tick label/.append style={font=\tiny}}

\newlength{\starsize}
\newlength{\starspread}
\tikzset{starsize/.code={\setlength{\starsize}{#1}},
         starspread/.code={\setlength{\starspread}{#1}}}
\tikzset{starsize=1mm,
         starspread=3mm}
\pgfdeclarepatternformonly[\starspread,\starsize]
  {my fivepointed stars}
  {\pgfpointorigin}
  {\pgfqpoint{\starspread}{\starspread}}
  {\pgfqpoint{\starspread}{\starspread}}
  {
   \pgftransformshift{\pgfqpoint{\starsize}{\starsize}}
   \pgfpathmoveto{\pgfqpointpolar{18}{\starsize}}
   \pgfpathlineto{\pgfqpointpolar{162}{\starsize}}
   \pgfpathlineto{\pgfqpointpolar{306}{\starsize}}
   \pgfpathlineto{\pgfqpointpolar{90}{\starsize}}
   \pgfpathlineto{\pgfqpointpolar{234}{\starsize}}
   \pgfpathclose%
   \pgfusepath{fill}
  }

\input{tex/macro}

\AtBeginDocument{%
  \providecommand\BibTeX{{%
    \normalfont B\kern-0.5em{\scshape i\kern-0.25em b}\kern-0.8em\TeX}}}

\setcopyright{acmcopyright}
\copyrightyear{2018}
\acmYear{2018}
\acmDOI{XXXXXXX.XXXXXXX}

\acmPrice{15.00}
\acmISBN{978-1-4503-XXXX-X/18/06}

\begin{document}


\title{Rethinking Semi-Supervised Node Classification with Self-Supervised Graph Clustering}


\author{Songbo Wang}
\authornote{Work done while being an intern at HKBU.}
\authornote{Both authors contributed equally to the paper.}
\affiliation{%
  \institution{The University of Hong Kong}
  \country{Hong Kong SAR, China}
}
\email{u3637109@connect.hku.hk}

\author{Renchi Yang}
\authornotemark[2]
\affiliation{%
  \institution{Hong Kong Baptist University}
  \country{Hong Kong SAR, China}
}
\email{renchi@hkbu.edu.hk}

\author{Yurui Lai}
\affiliation{%
  \institution{Hong Kong Baptist University}
  \country{Hong Kong, China}
}
\email{csyrlai@comp.hkbu.edu.hk}

\author{Xiaoyang Lin}
\affiliation{%
  \institution{Hong Kong Baptist University}
  \country{Hong Kong SAR, China}
}
\email{csxylin@comp.hkbu.edu.hk}

\author{Tsz Nam Chan}
\affiliation{%
  \institution{Shenzhen University}
  \city{Shenzhen}
  \country{China}
}
\email{edisonchan@szu.edu.cn}

\renewcommand{\shortauthors}{Wang and Yang, et al.}

\begin{abstract}
The emergence of {\em graph neural networks} (GNNs) has offered a powerful tool for semi-supervised node classification tasks. 
Subsequent studies have achieved further improvements through refining the message passing schemes in GNN models or exploiting various data augmentation techniques to mitigate limited supervision.
In real graphs, nodes often tend to form tightly-knit communities/clusters, which embody abundant signals for compensating label scarcity in semi-supervised node classification but are not explored in prior methods.

Inspired by this, this paper presents \algo{} that integrates self-supervised graph clustering and semi-supervised classification into a unified framework.
Firstly, we theoretically unify the optimization objectives of GNNs and spectral graph clustering, and based on that, develop {\em soft orthogonal GNNs} (SOGNs) that leverage a refined message passing paradigm to generate node representations for both classification and clustering. 
On top of that, \algo{} includes a self-supervised graph clustering module that enables the training of SOGNs for learning representations of unlabeled nodes in a self-supervised manner.
Particularly, this component comprises two non-trivial clustering objectives and a {\em Sinkhorn-Knopp} normalization that transforms predicted cluster assignments into balanced soft pseudo-labels.
Through combining the foregoing clustering module with the classification model using a multi-task objective containing the supervised classification loss on labeled data and self-supervised clustering loss on unlabeled data, \algo{} promotes synergy between them and achieves enhanced model capacity.
Our extensive experiments showcase that the proposed \algo{} framework consistently and considerably outperforms popular GNN models and recent baselines for semi-supervised node classification on seven real graphs, when working with various classic GNN backbones. 
\end{abstract}




\maketitle

\input{tex/introduction}

\input{tex/relatedwork}

\input{tex/preliminary}

\input{tex/method}

\input{tex/experiments}

\section{Conclusion}
In this paper, we propose \algo{}, a novel semi-supervised node classification framework that unifies self-supervised graph clustering with MPNNs through a soft orthogonal constraint. \algo{} combines self-supervised clustering with SKN to learn improved node representations, consistently outperforming state-of-the-art baselines. Specifically, \algo{} (i) aligns spectral clustering assignments with MPNN message passing for cluster-centered assignment optimization, enhancing meaningful unsupervised signals; and (ii) adapts to various GNN architectures and graph scales. Extensive experiments show that \algo{} achieves superior accuracy while exhibiting strong generalization and scalability, making it effective for semi-supervised node classification under limited supervision.

\begin{acks}
This work is partially supported by the National Natural Science Foundation of China (No. 62302414), the Hong Kong RGC ECS grant (No. 22202623) and YCRG (No. C2003-23Y), the Huawei Gift Fund, and Guangdong and Hong Kong Universities ``1+1+1'' Joint Research Collaboration Scheme, project No.: 2025A0505000002.
\end{acks}

\balance


\section*{Ethical Considerations}
The direct negative societal impacts of this research—specifically with respect to fairness, privacy, and security—are minimal. Nonetheless, as with other node classification models, erroneous predictions by the model may affect system functionality. Although the model's effectiveness has been extensively validated through experiments, occasional inaccuracies, particularly when processing noisy data, are still possible. To mitigate these risks, it is recommended to enhance data quality through rigorous data cleaning and preprocessing prior to model deployment.

\bibliographystyle{ACM-Reference-Format}
\bibliography{sample-base}

\appendix
\input{tex/appendix}

\end{document}

%% file: tex/macro.tex
\newcommand{\renchi}[1]{{\color{red}{[Renchi: #1]}}}

\newcommand{\songbo}[1]{{\color{blue}{[Songbo: #1]}}}

\makeatletter
\newcommand*\bigcdot{\mathpalette\bigcdot@{.5}}
\newcommand*\bigcdot@[2]{\mathbin{\vcenter{\hbox{\scalebox{#2}{$\m@th#1\bullet$}}}}}
\makeatother

\newcommand{\stitle}[1]{\vspace*{0.5em}\noindent{\bf #1.\/}}

\newcommand{\V}{\mathcal{V}\xspace}
\newcommand{\G}{\mathcal{G}\xspace}

\newcommand{\N}{\mathcal{N}\xspace}
\newcommand{\EDG}{\mathcal{E}\xspace}

\newcommand{\C}{\mathcal{C}\xspace}
\newcommand{\Y}{\mathcal{Y}\xspace}

\newcommand{\WM}{\mathbf{W}\xspace}
\newcommand{\AM}{\mathbf{A}\xspace}
\newcommand{\DM}{\mathbf{D}\xspace}
\newcommand{\IM}{\mathbf{I}\xspace}

\newcommand{\PM}{\mathbf{P}\xspace}
\newcommand{\CM}{\mathbf{C}\xspace}
\newcommand{\YM}{\mathbf{Y}\xspace}
\newcommand{\XM}{\mathbf{X}\xspace}
\newcommand{\LM}{\mathbf{L}\xspace}

\newcommand{\HM}{\mathbf{H}\xspace}
\newcommand{\ZM}{\mathbf{Z}\xspace}

\newcommand{\NAM}{\tilde{\mathbf{A}}\xspace}
\newcommand{\NLM}{\tilde{\mathbf{L}}\xspace}

\newcommand{\QM}{\mathbf{Q}\xspace}

\newcommand{\algo}{\texttt{NCGC}\xspace}

\newcommand{\eat}[1]{}

\def\addlegendimage{\csname pgfplots@addlegendimage\endcsname}

\makeatletter
\newcommand\footnoteref[1]{\protected@xdef\@thefnmark{\ref{#1}}\@footnotemark}
\makeatother

\let\oldnl\nl
\newcommand{\nonl}{\renewcommand{\nl}{\let\nl\oldnl}}

\DeclareMathOperator{\Tr}{Tr}

\definecolor{myred}{HTML}{fd7f6f}
\definecolor{myred_new}{HTML}{D8D8D8}
\definecolor{myred_new2}{HTML}{D7191C}
\definecolor{myblue}{HTML}{7eb0d5}
\definecolor{mygreen}{HTML}{b2e061}
\definecolor{mypurple}{HTML}{bd7ebe}
\definecolor{myorange}{HTML}{ffb55a}
\definecolor{myyellow}{HTML}{ffee65}
\definecolor{mypurple2}{HTML}{beb9db}
\definecolor{mypink}{HTML}{fdcce5}
\definecolor{mycyan}{HTML}{8bd3c7}

\definecolor{myblue2}{HTML}{115f9a}
\definecolor{myred2}{HTML}{c23728}



%% file: tex/introduction.tex
\section{Introduction}
In the past decade, {\em graph neural networks} (GNNs)~\cite{kipf2017semisupervised} have emerged as go-to models for learning on graphs due to its remarkable capabilities to comprehend and handle complex graph structures. Most GNNs typically follow the {\em message passing} scheme~\cite{gilmer2017neural}, wherein the features of a node are iteratively updated by aggregating and transforming the features from its neighborhood.
Owing to this simple yet effective mechanism, GNNs have seen fruitful successes over a variety of applications, including 
recommender systems~\cite{ying2018graph,he2020lightgcn,borisyuk2024lignn},
bioinformatics~\cite{abramson2024accurate,zhang2021graph},
financial fraud detection~\cite{dou2020enhancing,rao2021xfraud}, and many others~\cite{lam2023learning,de2024personalized,wu2022graph,jiang2025community}.

Amid various downstream tasks of GNNs, semi-supervised node classification is one of the most fundamental problems, whose goal is to infer the labels for the majority of unlabeled nodes when given a partially labeled graph.
Due to the the labor intensive data annotation in real world~\cite{li2019learning}, acquiring sufficient amount of accurate labels for model training is impractical, which in turn, limits the effectiveness of supervised learning in semi-supervised node classification.
Although the emergence of \texttt{GCN}~\cite{kipf2017semisupervised} has set a new benchmark, this task still remains tenaciously challenging due to severe label scarcity, over-smoothing issues, representation collapse, class imbalance, etc., which motivate a series of subsequent studies~\cite{zhang2025normalize,ding2024divide}. 

Over the past few years, considerable research efforts~\cite{waikhom2023survey} have been invested towards enhancing GNNs' representation learning capabilities by designing elaborate message passing mechanisms~\cite{gasteiger2018predict,xu2018representation,xu2018how,zhu2021interpreting,wu2019simplifying,gasteiger2019diffusion,veličković2018graph,huang2023node,zheng2026rethink}, achieving deeper models~\cite{liu2020towards,chen2020simple,li2019deepgcns,rongdropedge}, alleviating over-smoothing/squashing~\cite{zhao2020pairnorm,zhou2020towards,huang2024how}, etc.
Even though these studies made different attempts to remedy the inherent defects of GNNs, the limited availability of data labels and training samples hinders them from making further inroads in semi-supervised node classification performance.
Instead of ameliorating the model architecture, another line of research focuses on exploring various data augmentation strategies~\cite{feng2020graph,verma2021graphmix,wang2020nodeaug,zhao2022autogda,lai2024efficient} to improve model generalization under limited supervision. A major category of data augmentation approaches constructs different augmented graphs by perturbing, removing, masking, corrupting, or sampling edges, nodes, features, or subgraphs in the input graph, thereby either implicitly creating additional labeled samples for model training, or providing augmented views to enable contrastive learning for more discriminative representations~\cite{yang2024mixed}.
Distinctly, the second class of augmentation methods~\cite{dong2021equivalence,ding2022meta,ding2024divide,sun2020multi,zhang2025normalize} resort to pseudo-labeling tricks~\cite{lee2013pseudo}, most of which typically create pseudo-labels using techniques like label propagation with labeled data, followed by using them as additional supervisory signals.

\begin{table}[!t]
\centering
\renewcommand{\arraystretch}{1.0}
\begin{small}
\caption{Clustering and classification accuracy results.}\label{tbl:clas-cluster}
\vspace{-3mm}
\begin{tabular}{l|cc|cc}
	\hline
	\multirow{2}{*}{\bf Dataset} & \multicolumn{2}{c|}{\bf \makecell{Unsupervised\\ Clustering}} & \multicolumn{2}{c}{\bf \makecell{Semi-supervised\\ Classification}}  \\
    & \texttt{S\textsuperscript{3}GC}~\cite{devvrit2022s3gc} & \texttt{MVGRL}~\cite{hassani2020contrastive} & \texttt{GCN}*~\cite{luoclassic} & \texttt{GAT}*~\cite{luoclassic} \\ \hline
    {\em Cora} &  74.2 & 76.3 & 84.60 &  83.82  \\
    {\em CiteSeer} & 68.8 & 70.3 & 72.42 &  70.94  \\
    {\em PubMed} & 71.3 & 67.5 & 80.52  &  79.06  \\
    \hline
\end{tabular}%
\end{small}
\vspace{-3ex}
\end{table}

Despite the progress made, the majority of the foregoing works still suffer from severe flaws, including noise introduced by augmentations or error accumulation in pseudo labels, which limit their model generality and robustness. 
More importantly, real-world graphs inherently embody rich community or cluster structures, where nodes within the same cluster often share similar labels, attributes, and topological semantics.
Particularly, as reported in Table~\ref{tbl:clas-cluster},
in the absence of node labels, state-of-the-art graph clustering approaches~\cite{devvrit2022s3gc,hassani2020contrastive} are able to attain decent accuracies on benchmark datasets~\cite{SenNBGGE08coracitesser} when measured against ground-truth class labels.
For instance, the performance gaps from fully-tuned \texttt{GCN}~\cite{luoclassic} on {\em CiteSeer} and {\em PubMed} are merely $2.12\%$ and $3.02\%$, respectively.
However, most existing semi-supervised classification models solely rely on a few node annotations for model training but rarely capitalize on such community or cluster information from unlabeled data. In other words, re-purposing and exploiting self-supervised clustering to guide semi-supervised node classification for enhanced performance still remains largely underexplored.

\stitle{Present Work} To bridge this gap, we propose \algo{} (short for semi-supervised \underline{N}ode \underline{C}lassification with self-supervised \underline{G}raph \underline{C}lustering), a unified framework that seamlessly integrates self-supervised graph clustering with semi-supervised classification in a multi-tasking strategy, and leverages the synergy between them for effective node classification.
Under the hood, \algo{} consists of (i) a theoretically-grounded model {\em soft orthogonal GNNs} (SOGNs), which is tailored to learning node representations for both graph clustering and node classification, and (ii) a well-thought-out self-supervised graph clustering module that effectively extracts supervision signals from unlabeled data.

Specifically, SOGNs are motivated by the theoretical connection between popular GNNs and {\em spectral graph clustering}~\cite{von2007tutorial}, upon which we propose to include the {\em soft orthogonal constraint} (SOC) into the optimization framework of GNNs, engendering an optimized message passing strategy for updating representations in both tasks.
Since it is built on the unified optimization framework of GNNs, SOGNs can work with any message passing operators in existing GNNs as the fundamental operations therein. 
Based thereon, we learn the representations of unlabeled nodes by training the SOGN model with optimizing {\em deep embedded clustering}~\cite{xie2016unsupervised} objective and a new pseudo-labeling clustering loss in a self-supervised fashion, where the former aims at matching the predicted node-cluster assignments of unlabeled nodes to the target {\em Student’s t-distribution}~\cite{maaten2008visualizing}, while the latter is to align node-cluster predictions with pseudo cluster labels created via {\em Sinkhorn-Knopp normalization} (SKN)~\cite{knight2008sinkhorn}.
By sharing the SOGNs as the backbone model between the semi-supervised classification with labeled nodes and self-supervised clustering with unlabeled nodes in the downstream phase, \algo{} enables them to benefit from each other for improved model capacity and mutually enhance discriminative abilities.

Our extensive experiments evaluating our \algo{} framework with four popular GNN backbones against 22 baselines on 7 real graph datasets demonstrate that \algo{} conspicuously and consistently outperforms classic GNN models and recent state-of-the-art competitors in terms of semi-supervised node classification. 

%% file: tex/relatedwork.tex
\section{Related Work}

\subsection{Semi-Supervised Node Classification}

Semi-supervised node classification is a core task in graph representation learning with applications in recommendation, bioinformatics, and fraud detection domains. Existing works can be broadly categorized into architecture design, data augmentation, and pseudo-labeling methods.

\stitle{Architecture design}
Early GNNs such as GCN~\cite{kipf2017semisupervised} and GraphSAGE~\cite{hamilton2017inductive} popularized message passing, whereas GAT~\cite{veličković2018graph} introduced attention to adaptively weight neighbors. To address over-smoothing, APPNP~\cite{gasteiger2018predict}, DAGNN~\cite{liu2020towards}, and GCNII~\cite{chen2020simple} decouple propagation and transformation or enhance them with residual/identity mappings, while JK-Net~\cite{xu2018representation} fuses layers for richer aggregation. In pursuit of efficiency, SGC~\cite{wu2019simplifying} collapses propagation into a single linear step. Extending structural modeling, Geom-GCN~\cite{pei2020GeomGCN} leverages latent geometry, and Bayesian-GCN~\cite{zhang2019bayesian} models weight uncertainty for robustness.

\stitle{Data augmentation}
Data augmentation techniques have been extensively explored to improve generalization in graph learning. For instance, GRAND~\cite{feng2020graph} creates stochastic propagated views for consistency, while GraphMix~\cite{verma2021graphmix} and NodeAug~\cite{wang2020nodeaug} diversify inputs via interpolation or corruption. GAM~\cite{stretcu2019graph} and Violin~\cite{xie2023violin} regularize predictions through agreement or virtual edges. Building on this idea, MGCN~\cite{yang2024mixed} couples augmentation with contrastive learning, and AutoGDA~\cite{zhao2022autogda} leverages automated search to identify optimal augmentation strategies. Further studies~\cite{park2021metropolis,tang2021data,zhao2021data} integrate edge perturbation, feature masking, and subgraph sampling to jointly enhance diversity and robustness.

\stitle{Pseudo-labeling methods}
Pseudo labeling~\cite{lee2013pseudo} enables using unlabeled nodes. Meta-PN~\cite{ding2022meta} formulates label propagation as meta-learning for few-shot generalization. M3S~\cite{sun2020multi} refines labels in stages, whereas DND-Net~\cite{ding2024divide} denoises them. PTA~\cite{dong2021equivalence} and NormProp~\cite{zhang2025normalize} stabilize pseudo supervision by coupling propagation with feature learning. Complementary efforts~\cite{ding2024toward, li2018deeper, wang2023few} address low-resource and weak supervision via confidence thresholds and curriculum strategies.

\subsection{Self-Supervised Graph Clustering}
Compared to traditional unsupervised clustering, self-supervised clustering leverages auxiliary tasks, such as pseudo-labeling, data augmentation, contrastive learning, reconstruction, or attribute prediction, to learn more discriminative representations and improve clusterability. Recent works design various forms of pretext or supervision signals to guide representation learning, enhance separation, and improve generalization across datasets.

\stitle{Pretext \& Pseudo-Label Methods} 
Early methods design pretext tasks such as structure reconstruction, attribute prediction, or mutual information maximization. For instance, S3GC~\cite{devvrit2022s3gc} combines spectral embedding with self-supervised training to improve separation, while DSAGC~\cite{lu2024deep} enhances pseudo-label reliability through high-confidence sample selection. ENID~\cite{zhu2024every} dynamically assigns node-specific tasks based on local structure and feature complexity. Other works refine pseudo-label usage via reliable label selection~\cite{lu2025novel}, prompt-based guidance~\cite{xia2022self}, or ensemble refinement~\cite{zhu2022collaborative}, aiming to stabilize training and mitigate noise.

\stitle{Contrastive Learning-based Clustering} 
In parallel, contrastive learning aligns node embeddings across multiple graph views while separating different clusters. Representative approaches include SCGC~\cite{kulatilleke2025scgc} with a simplified loss, GraphLearner~\cite{yang2024graphlearner} with learnable view generation, and DCC~\cite{peng2023dual} combining instance- and cluster-level objectives. Subsequent works enhance these approaches via adaptive filtering~\cite{xie2023contrastive}, explicit community modeling~\cite{park2022cgc}, or multi-view/multi-layer designs~\cite{xia2021self, liu2022multilayer}, often achieving strong results even without explicit pseudo-labeling or reconstruction losses.

\subsection{Clustering-based Self-Supervised Learning}
While limited efforts have been devoted to node classification in this direction, clustering-based self-supervised methods have achieved remarkable success in computer vision by discretizing the feature space and leveraging cluster assignments as supervisory signals. For instance, DeepCluster v2~\cite{caron2018deep} performs offline $k$-means clustering to generate pseudo-labels for training. To prevent degenerate solutions, SeLa~\cite{asano2019self} introduces uniform constraints to encourage balanced cluster assignments. Building upon this, PCL~\cite{li2020prototypical} enhances cluster stability through memory-based consistency across training iterations. Furthermore, online approaches such as SwAV~\cite{caron2020unsupervised} compute cluster assignments within mini-batches via {\em Sinkhorn-Knopp normalization}, enforcing consistency between augmented views using cross-entropy loss. DINO~\cite{caron2021emerging}, on the other hand, adopts a teacher-student framework to align soft predictions across views without relying on explicit clustering.

These methods effectively convert the unsupervised learning problem into a supervised-like task by aligning predictions with discrete or soft targets, thereby introducing structured semantic supervision. However, such clustering-based objectives have been largely overlooked in graph representation learning, particularly under the semi-supervised settings. Our work aims to bridge this gap by aligning graph representation learning with spectral clustering through soft orthogonal constraints, offering both interpretability and an effective training signal for GNNs.

%% file: tex/preliminary.tex
\section{Preliminaries}

\subsection{Graph Nomenclature}

Let $\G=(\V,\EDG,\XM)$ be an attributed graph, where $\V$ is a set of $n$ nodes and $\EDG$ is a set of $m$ edges. For each edge $(v_i, v_i)\in \EDG$, we say $v_i$ and $v_j$ are neighbors to each other, and we use $\N(v_i)$ to denote the set of neighbors of $v_i$, with the degree $d(v_i)=|\N(v_i)|$. $\XM\in\mathbb{R}^{n\times D}$ is the input attribute matrix of nodes, where each row $\XM_i$ stands for the attributes associated with node $v_i$.
We use $\AM$ to symbolize the adjacency matrix of $\G$, where $\AM_{i,j}=1$ if there is an edge $(v_i, v_j)\in\EDG$, and otherwise $\AM_{i,j}=0$, and $\DM$ to denote the degree matrix of $\G$. Accordingly, $\LM=\DM-\AM$ is used to represent the Laplacian, and $\PM$ denotes the transition matrix of $\G$, respectively. $\NAM=\DM^{-\frac{1}{2}}\AM\DM^{-\frac{1}{2}}$ and $\NLM=\IM-\NAM$ are the normalized adjacency and Laplacian matrices of $\G$, respectively. 

\stitle{Semi-supervised Node Classification} Given a graph $\G$, $K$ distinct class labels, and partially observed labels $\Y_{l}=\{y_1,y_2,\ldots,y_{|\V_{{l}}|}\}$ for nodes $\V_{{l}}\subseteq \V$, the semi-supervised node classification aims to predict the class labels for unlabeled nodes $\V_{{u}}=\V\setminus \V_{{l}}$ with $|\V_{{l}}|\ll |\V_{{u}}|$.
For ease of exposition, we denote by $\YM\in \mathbb{R}^{n\times K}$ the ground-truth class labels for all nodes in $\V$, where $\YM_i=\{0,1\}^K$ stands for the one-hot label vector of node $v_i$.

\begin{table}[t]
\centering
\caption{$\HM, \HM^{(0)}, \lambda$, and $\Omega$ in various GNNs.}\label{tbl:GNNs}
\vspace{-2ex}
\resizebox{\columnwidth}{!}{
\begin{tabular}{llccc}
\toprule
{\bf Model} & \(\HM\) & \(\lambda\) & \(\Omega\) & \(\HM^{(0)}\)\\
\midrule
GCN/SGC~\cite{kipf2017semisupervised} & \(\HM=\NAM^T\HM^{(0)}\) & \(1\) &  - & \(\texttt{LN}(\XM)\)\\
PPNP~\cite{gasteiger2019combining} & \(\HM=\alpha\left(\IM -(1-\alpha)\NAM \right)^{-1}\HM^{(0)}\) & \(\frac{1}{\alpha}-1\) &  \(\|\HM-\HM^{(0)}\|_\text{F}^2\) &\(\texttt{MLP}(\XM)\)  \\
APPNP~\cite{gasteiger2019combining} & \(\HM=\alpha\sum_{t=0}^T{(1-\alpha)^t\NAM^t\HM^{(0)}}\) & \(\frac{1}{\alpha}-1\) &  \(\|\HM-\HM^{(0)}\|_\text{F}^2\) &\(\texttt{MLP}(\XM)\)  \\
JKNet~\cite{xu2018representation} & \(\HM=\sum_{k=1}^T{\frac{\alpha^{t-1}}{(1+\alpha)^{t}}\NAM^t\HM^{(0)}}\) & \(\alpha\) & \(\|\NAM\HM-\HM^{(0)}\|^2_\text{F}\) & \(\texttt{LN}(\XM)\) \\
DAGNN~\cite{liu2020towards} & \(\HM=\sum_{t=0}^T{\frac{\alpha^{t}}{(1+\alpha)^{t+1}}\NAM^t\HM^{(0)}}\) & \(\alpha\) & \(\|\HM-\HM^{(0)}\|_F^2\) & \(\texttt{MLP}(\XM)\) \\
\bottomrule
\end{tabular}
}
\vspace{-2ex}
\end{table}

\subsection{Graph Neural Networks}\label{sec:GNNs}
Most GNNs are multi-layer feedforward neural networks, which comprise two layer-wise operators: a feature transformation function $\texttt{Trans}$ (usually an MLP) and an aggregation function $\texttt{Aggr}$ that follows the message passing scheme~\cite{gilmer2017neural}. Denote by $\HM^{(t+1)}_i\ \in \mathbb{R}^d$ the hidden representation of node $v_i$ at the $t$-th layer and $d$ is the dimension of hidden representations. Particularly, $\HM^{(0)} = f(\XM)$, where $f(\cdot)$ can be an identity mapping, linear transformation, or MLP.
The output node representation at the $(t+1)$-th layer can be formulated as\footnote{$\NAM$ can also be replaced by $\PM$.}
\begin{equation}
\HM^{(t+1)}_i = \texttt{Trans}\left(\HM^{(t)}_i, \texttt{Aggr}\left(\sum_{v_j\in \N(v_i)}\NAM_{i,j}\HM^{(t)}_j\right)\right).
\end{equation}
For the classic GCN~\cite{kipf2017semisupervised}, the updating rule can be written as
\begin{equation}\label{eq:GCN-layer}
\HM^{(t+1)} = \sigma(\NAM\HM^{(t)}\WM^{(t)}),
\end{equation}
where $\WM^{(t)}$ is trainable weight matrix at the $t$-th layer, $\sigma(\cdot)$ stands for the activation function such as ReLU. 

As demystified in recent studies~\cite{ma2021unified,yang2021attributes, zhu2021interpreting}, after removing the non-linearity, many existing popular GNNs~\cite{Kipf17gcn, Wu19sgc,gasteiger2018predict, gasteiger2019diffusion,chen2020simple,xu2018representation} can be unified into a {\em graph Laplacian smoothing}~\cite{dong2016learning} problem as follows:
\begin{align}
\label{eq:GNN_obj}    
    \min_{\HM\in \mathbb{R}^{n\times d}} \lambda \cdot\Tr(\HM^{\top} \NLM\HM ) + \Omega,
\end{align}
where $\HM$ denotes the target node representations and $\lambda$ stands for a coefficient striking a balance between the two terms.
The second term aims to reduce the discrepancy between the initial node features and $\HM$, while the first term can be rewritten as $\textstyle \lambda\cdot\sum_{(v_i, v_i)\in\EDG}\left\| \frac{\HM_i}{\sqrt{d(v_i)}}- \frac{\HM_j}{\sqrt{d(v_j)}}\right\|^2$, meaning that the node features of adjacent nodes are enforced to be similar.
Table~\ref{tbl:GNNs} lists the formulations of $\HM$, $\HM^{(0)}$, $\lambda$, and $\Omega$ in various GNNs that optimize Eq.~\eqref{eq:GNN_obj}.

\subsection{Spectral Graph Clustering}
\eat{
As in common practice, the clustering result can be represented by a {\em Normalized Cluster Indicator} (NCI) $\YM\in \mathbb{R}^{n\times K}$ defined as follows:
\begin{equation}\label{eq:NCI}
\CM_{i,j} = \begin{cases}
\frac{1}{\sqrt{|\mathcal{C}_j|}}, & \text{if $v_i\in \mathcal{C}_j$}, \\
\quad 0, & \text{otherwise}.
\end{cases}
\end{equation}
In particular, the columns of $\CM$ are orthonormal, i.e., $\CM^{\top}\CM=\IM$.
}

{\em Spectral clustering}~\cite{von2007tutorial} is a canonical technique for graph clustering, which seeks to partition nodes in $\G$ into disjoint groups $\{\C_1,\ldots,\C_K\}$ such that the inter-cluster connectivity is minimized.
One standard formulation of such objectives is the RatioCut~\cite{hagen1992new}: 
\begin{equation*}
\min_{\{\C_1,\ldots,\C_K\}}\sum_{k=1}^{K}{\frac{1}{K}\sum_{v_i\in \C_k, v_j\in \V\setminus \C_k}{\frac{{\NAM}_{i,j}}{|\C_k|}}}.
\end{equation*}
Intuitively, it is to minimize the average weight of edges connecting nodes in any two distinct clusters. As analysed in \cite{von2007tutorial,xie2025diffusion}, the above objective is equivalent to finding a cluster indicator matrix $\CM \in \mathbb{R}^{n\times K}$ optimizing the following trace minimization problem:
\begin{equation}\label{eq:spectral-loss}
\min_{\CM}\Tr(\CM^{\top}\NLM\CM),
\end{equation}
where $\forall{v_i\in \V}$ and $\forall{1\le k\le K}$,
\begin{equation}\label{eq:NCI}
\CM_{i,k} = \begin{cases}
\frac{1}{\sqrt{|\mathcal{C}_k|}}, & \text{if $v_i\in \mathcal{C}_k$}, \\
\quad 0, & \text{otherwise},
\end{cases}
\end{equation}
and $\CM$ is column-orthonormal, i.e., $\CM^\top\CM=\IM$.

The optimization objective in Eq.~\eqref{eq:spectral-loss} is an NP-hard problem given the constraint on $\CM$ as stated in Eq.~\eqref{eq:NCI}. A common way is to compute an approximate solution by relaxing the discreteness condition on $\CM$ and allowing it to take arbitrary values in $\mathbb{R}$ such that the column-orthonormal property, i.e., $\CM^{\top}\CM=\IM$, still holds. By Ky Fan’s trace maximization principle~\cite{fan1949theorem}, it immediately leads to that the optimal solution is the $k$-largest eigenvectors $\QM$ of ${\NAM}$. The $K$-Means or {\em rounding algorithms}~\cite{shi2003multiclass,yang2024efficient} are then applied to convert the $\QM$ into the cluster indicator $\CM$ that satisfies Eq.~\eqref{eq:NCI}.

\subsection{Connecting GNNs and Spectral Clustering}\label{sec:GNN-SC}

According to the optimization objective of spectral clustering in Eq.~\eqref{eq:spectral-loss}, it can also be perceived as a graph Laplacian smoothing problem with an additional constraint on the variable $\CM$ stated in Eq.~\eqref{eq:NCI}. In other words, spectral clustering is a variant of GNNs without learning weights and non-linear operations, which aims at optimizing Eq.~\eqref{eq:GNN_obj} with $\Omega=0$ and the column-orthonormal requirement on $\HM$, i.e., $\HM^\top\HM=\IM$. 

\begin{algorithm}[h]
\caption{Subspace Iteration Method~\cite{saad2011numerical}}\label{alg:eigh}
\begin{algorithmic}[1]
\State \textbf{Start:} Choose an initial system of vectors $\QM^{(0)}\in \mathbb{R}^{n\times K}$\;
\State{\textbf{Iterate:} Until convergence do,
\State{\quad\quad\quad\ \ $\tilde{\QM}^{(t-1)} \gets \NAM\QM^{(t-1)}$} 
\Comment{Propagation Step}
\State{\quad\quad\quad\ \ $\QM^{(t)} \gets \texttt{QR}(\tilde{\QM}^{(t-1)})$}
\Comment{Orthonormalization Step}
}
\end{algorithmic}
\end{algorithm}

\begin{algorithm}[h]
\caption{Message Passing Scheme~\cite{kipf2017semisupervised}}\label{alg:MP}
\begin{algorithmic}[1]
\State \textbf{Start:} Initialize $\HM^{(0)}=f(\XM)$\;
\State{\textbf{Iterate:} $t\gets 1$ to $T$ do,
\State{\quad\quad\quad\ \ $\tilde{\HM}^{(t-1)} \gets \NAM\HM^{(t-1)}$} 
\Comment{Propagation Step}
\State{\quad\quad\quad\ \ $\HM^{(t)} \gets \texttt{Trans}(\tilde{\HM}^{(t-1)})$}
\Comment{Transformation Step}
}
\end{algorithmic}
\end{algorithm}

More specifically, Algorithms~\ref{alg:eigh} and ~\ref{alg:MP} present the pseudo-code of the partial eigendecomposition and message passing method used in spectral clustering and GNNs towards solving their respective optimization objectives. Note that we utilize the canonical {\em subspace iterations}~\cite{saad2011numerical} for partial eigendecomposition as an example for illustration as popular methods, e.g., power iterations and Arnoldi iterations follow a similar workflow. As outlined in Algorithms~\ref{alg:eigh} and ~\ref{alg:MP}, both methods can be summarized into iterative procedures wherein the first step is propagating the temporary solutions $\QM^{(t-1)}$ and $\HM^{(t-1)}$ with the normalized adjacency matrix $\NAM$, followed by an orthonormalization using QR decomposition and a transformation using a linear network or MLP, respectively. Another difference lines on the initialization, during which Algorithm~\ref{alg:eigh} randomly choose a system of vectors of $\QM^{(0)}$ such that $\QM^{(0)\top}\QM^{(0)}=\IM$, while Algorithm~\ref{alg:MP} converts input node attribute vectors $\XM$ into $\HM^{(0)}$ using a function $f(\cdot)$.

Moreover, in their respective downstream tasks, i.e., node clustering and classification, spectral clustering converts $\QM$ into the $n\times K$ node-cluster indicator $\CM$, while GNNs map $\HM$ into the $n\times K$ node-class predictions $\YM^\prime$ using a linear layer (or MLP) as well as the softmax function.

%% file: tex/method.tex
\section{Methodology}
In this section, we present our \algo{} framework integrating semi-supervised graph clustering for semi-supervised node classification. We first provide an overview of \algo{} in \S~\ref{sec:overview}, followed by elaborating on our backbone module {\em Soft Orthogonal GNNs} (SOGNs) in \S~\ref{sec:model-layers}. 
In~\S~\ref{sec:SSC}, we delineate how to realize self-supervised clustering based on our backbone and unearth self-supervision signals for unlabeled nodes.

\subsection{Framework Overview}\label{sec:overview}

\begin{figure*}[!t]
    \centering
    \includegraphics[width=0.95\textwidth]{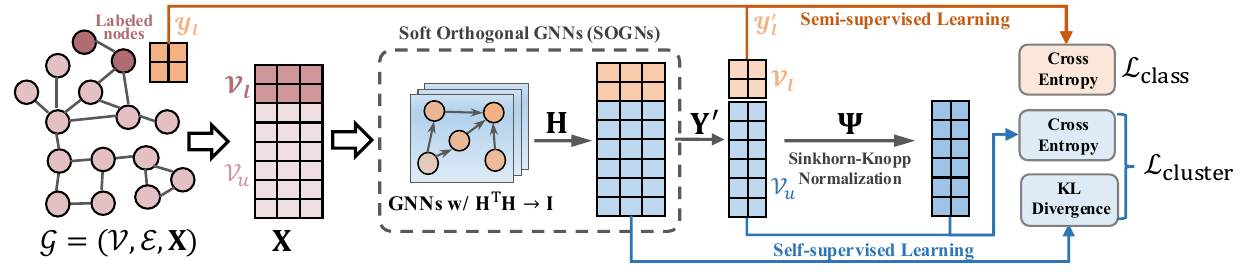}
    \vspace{-2ex}
    \caption{Overview of \algo{}.}
    \label{fig:overview}
\end{figure*}

At a high level, the idea of \algo{} is to leverage the complementary nature of semi-supervised learning with labeled nodes and self-supervised learning with unlabeled nodes, and hence, achieve synergistic optimizations of their respective class predictions and cluster assignments.
Given the close connections between classification and clustering, \algo{} unifies these two tasks into a single framework via a multi-tasking strategy.

As illustrated in Figure~\ref{fig:overview}, given input graph $\G$ with partial node set $\V_l$ that is associated with ground-truth node labels $\Y_l$, and unlabeled node set $\V_u$, \algo{} first transforms input attribute vectors $\XM$ of all nodes into node representations $\HM$ using our backbone model SOGNs.
Subsequently, \algo{} learns node representations $\HM$ by simultaneously training the model in the semi-supervised and self-supervised settings. More specifically, \algo{} computes the class/cluster predictions via a bias-free linear layer parameterized by $\WM_{\text{proto}}\in \mathbb{R}^{d\times K}$ and followed by a softmax function:
\begin{equation}\label{eq:Yprime}
\YM^\prime = \textsf{softmax}(\HM\WM_{\text{proto}}).
\end{equation}
Based thereon, we inject semi-supervision signals into the model training through optimizing the supervised objective, i.e., cross-entropy loss, that evaluates class predictions $\YM^\prime[\V_l]$ for labeled nodes $\V_l$ against the ground-truth labels in $\Y_l$:
\begin{equation}
\mathcal{L}_{\text{class}} = -\sum_{v_i\in \V_l}\sum_{k=1}^K{\YM^\prime_{i,k}\cdot \log{\YM_{i,k}}}.
\end{equation}
As per Eq.~\eqref{eq:Yprime}, $\WM_{\text{proto}}$ can be perceived as the learnable prototypes of $K$ classes when training the model solely with node labels as aforementioned.

In the meantime, to exploit and extract useful features underlying unlabeled data pertinent to $\V_u$, \algo{} further trains the model by utilizing $\HM$ and derived cluster predictions/assignments for self-supervised clustering tasks. More formally, this focuses on optimizing a joint clustering loss $\mathcal{L}_{\text{cluster}}$ upon unlabeled nodes in $\V_u$:
\begin{equation}
\mathcal{L}_{\text{cluster}} = \mathcal{L}_{\text{KL}} + \mathcal{L}_{\text{PL}}.
\end{equation}
Therein, $\mathcal{L}_{\text{KL}}$ is a deep embedded clustering objective minimizing the Kullback–Leibler (KL) divergence between the cluster assignments derived from $\HM$ for $\V_u$ and a pre-defined target distribution, while $\mathcal{L}_{\text{PL}}$ converts cluster predictions of $\V_u$ in $\YM^\prime$, i.e., $\YM^\prime[\V_u]$, into pseudo-labels for supervised learning.

Through sharing the backbone model SOGNs for classification and clustering tasks, \algo{} is able to benefit from the semi-supervision and self-supervision signals from both of them, and improve the model capacity. In particular, by leveraging $\WM_{\text{proto}}$ for producing both class predictions for labeled nodes and cluster assignments for unlabeled nodes, \algo{} essentially aligns class prototypes with cluster centroids, enabling them to mutually enhance each other's discriminative abilities.

\subsection{Soft Orthogonal GNNs}\label{sec:model-layers}

Recall that in \S~\ref{sec:GNN-SC}, we uncover that both GNNs for node classification and spectral graph clustering are built upon graph Laplacian smoothing, and meanwhile, share similar workflows involving propagation with $\NAM$, transformation/orthonormalization, and generating node-cluster/-class predictions. Based on this finding, we propose SOGNs that unify spectral clustering and GNNs to learn node representations $\HM$ for bolstering both clustering and classification tasks in the downstream phase. 

To realize the above-said idea, instead of adhering to the hard orthonormal constraint in classic spectral clustering, SOGNs includes a {\em soft orthogonal constraint} (SOC) in the graph Laplacian smoothing-based objective for learning node representations. Formally,
\begin{align}
\label{eq:GNN_obj_new}    
\min_{\HM\in \mathbb{R}^{n\times d}} \Tr(\HM^{\top} \NLM\HM ) + \beta\cdot \|\HM^\top\HM-\IM\|^2_{\text{F}},
\end{align}
where $\beta$ signifies the coefficient for adjusting the strength of the orthogonal constraint, which is typically set to $0.005$.
By setting the derivative of the above objective function w.r.t. $\HM$ to zero, we can obtain the optimal solution to $\HM$ as: 
\begin{align*}
& \frac{\partial \Tr(\HM^{\top} \NLM\HM )+ \beta\cdot\|\HM^\top\HM-\IM\|^2_{\text{F}}}{\partial \HM} = 0,
\end{align*}
which further leads to
\begin{equation}
 \begin{gathered}
\NLM\HM + \beta\cdot\HM \left(\HM^\top\HM-\IM\right) = 0 \\
\Rightarrow \HM = \NAM\HM - {\beta}\cdot \HM\HM^\top\HM.
\end{gathered}
\end{equation}
We resort to an iterative form to approximate the converged solution of $\HM$, namely,
\begin{equation*}
\HM^{(\ell+1)} = \NAM\HM^{(\ell)} - \beta\cdot \tilde{\HM}^{(\ell)}\tilde{\HM}^{(\ell)\top}\HM^{(\ell)}.
\end{equation*}

Following classic GNN operations in Eq.~\eqref{eq:GCN-layer}, we additional incorporate learnable weights $\WM^{(\ell)}$ and non-linear operator (i.e., activation function) $\sigma(\cdot)$. The node representations at the $\ell+1$-th layer are then computed as follows:
\begin{equation}\label{eq:update-H-new}
\begin{gathered}
\HM^{(\ell+1)} = \sigma\left(\NAM\ZM^{(\ell)} - \beta\cdot \tilde{\ZM}^{(\ell)}\tilde{\ZM}^{(\ell)\top}\ZM^{(\ell)}\right),\\
\text{where } \ZM^{(\ell)} = \HM^{(\ell)}\WM^{(\ell)}.
\end{gathered}
\end{equation}
$\tilde{\ZM}^{(\ell)}$ stands for the column-wise $L_2$ normalization of ${\ZM}^{(\ell)}$. This normalization is to avoid gradient exploding~\cite{guo2022orthogonal} when the number of layers in SOGNs increases. 
The initial node representations $\HM^{(0)}$ are transformed from the input attribute vectors via a linear layer or an MLP $f(\XM)$ with an activation as in previous GNNs.
Compared to GNN models revisited in \S~\ref{sec:GNNs}, the only difference in Eq.~\eqref{eq:update-H-new} lies on subtracting the term $\beta\cdot \tilde{\HM}^{(\ell)}\tilde{\HM}^{(\ell)\top}\HM^{(\ell)}$, which is to impose the SOC to the output embeddings $\HM$, thereby making $\HM$ suitable for clustering tasks.
It is worth mentioning that SOGNs can work with any message-passing-based GNN backbones (e.g., \texttt{GCN}, \texttt{GAT}, \texttt{GraphSAGE}), simply by replacing the propagation step $\NAM\HM^{(\ell)}$ in Eq.~\eqref{eq:update-H-new} by the corresponding message passing operator in them.

\subsection{Self-Supervised Graph  Clustering}\label{sec:SSC}
In what follows, we adapt SOGNs for self-supervised graph clustering. With the node representations $\HM$ output by SOGNs at hand, the clustering module of \algo{} is trained by optimizing an objective for deep embedded clustering~\cite{xie2016unsupervised} and another pseudo-labeling objective~\cite{caron2020unsupervised}. The former learns cluster centroids and matches the the predicted soft node-cluster assignments to an auxiliary target distribution unsupervisedly, while the latter seeks to align node-cluster predictions based on class prototypes with pseudo cluster labels in a self-supervised manner.

\stitle{Deep Embedded Clustering}
We formulate the first objective as a KL divergence between the soft assignments $P$ and the target distribution $Q$ as follows:
\begin{equation}\label{loss:KL}
\mathcal{L}_{\textnormal{KL}} = \texttt{KL}(P\|Q) = \sum_{v_i\in \V}\sum_{k=1}^{K}{P_{i,k}\cdot \log{\frac{P_{i,k}}{Q_{i,k}}}},
\end{equation}
where $Q_{i,k}$ represents the soft assignment, i.e., the probability of assigning node $v_i$ to cluster $\C_k$, and $P_{i,k}$ denotes the soft probabilistic target of $Q_{i,k}$.
Following~\cite{xie2016unsupervised}, the Student's t-distribution by \citet{maaten2008visualizing} is employed as a kernel to measure the similarity between each node and the cluster, i.e., soft assignment:
\begin{equation}\label{eq:q}
Q_{i,k} = \frac{(1+\|\HM_i - \CM_k\|^2)^{-1}}{\sum_{\ell=1}^K(1+\|\HM_i - \CM_\ell\|^2)^{-1}},
\end{equation}
where the rows in $\CM$ are $K$ learnable cluster centroids. As for the target distribution $P$, we calculate $P_{i,k}$ by first raising $Q_{i,k}$ to the second power and then normalizing it by frequency per cluster:
\begin{equation}
P_{i,k} = \frac{Q_{i,k}^2/\sum_{v_j\in\V}{Q_{j,k}}}{\sum_{\ell=1}^K{Q_{i,\ell}^2/\sum_{v_j\in\V}{Q_{j,\ell}}}}.
\end{equation}
Such a target distribution is softer and able to (i) improve cluster purity, (ii) emphasize nodes assigned with high confidence, and (iii) normalize loss contribution of each centroid to prevent large clusters from distorting the hidden feature space~\cite{xie2016unsupervised}.


\stitle{Pseudo-labeling Clustering}
As displayed in Eq.~\eqref{eq:PL}, our second objective adopts the cross-entropy loss to contrast targets (cluster assignments or labels) $\boldsymbol{\Psi}$ and node-cluster predictions $\boldsymbol{\Psi}^\prime$ of unlabeled nodes in $\V_u$ as in the standard supervised learning objective.
\begin{equation}\label{eq:PL}
\mathcal{L}_{\text{PL}} = -\sum_{v_i\in \V_u}\sum_{k=1}^K{\boldsymbol{\Psi}^\prime_{i,k}\cdot \log{\boldsymbol{\Psi}_{i,k}}}
\end{equation}    
By creating targets from model predictions, this pseudo-labeling paradigm transforms unsupervised data into a supervised learning problem, enabling the model to bootstrap its own supervision and driving the learning of more discriminative features, and gradually refining the representations, even in the absence of ground-truth labels.

Instead of contrasting predicted cluster assignments of correlated views of the same instance as in previous pseudo-labeling works~\cite{caron2020unsupervised,caron2021emerging} for images, \algo{} directly calculates the
soft node-cluster predictions $\boldsymbol{\Psi}^\prime$ for unlabeled nodes $\V_u$ by
\begin{equation}
\boldsymbol{\Psi}^\prime = \YM^\prime[\V_u] = \textsf{softmax}(\HM\WM_{\text{proto}})[\V_u],
\end{equation}
and then derive target cluster assignments $\boldsymbol{\Psi}$ therefrom. 
Note that we project unlabeled nodes in $\V_u$ to $K$ distinct clusters using the parameters $\WM_{\text{proto}}\in \mathbb{R}^{d\times K}$ shared with labeled nodes in Eq.~\eqref{eq:Yprime}. 
Since the columns in $\WM_{\text{proto}}$ can be regarded as $K$ cluster centroids here, this trick enables us to essentially align cluster centroids with class prototypes learned with the semi-supervision signals~\cite{fini2023semi}.



Next, we transform $\boldsymbol{\Psi}^\prime$ into pseudo-labels $\boldsymbol{\Psi}$ in the form of soft cluster assignments rather than hard assignments.
The reason is that using hard labels is more likely to make the model overfit to noisy or incorrect pseudo-labels.
In contrast, the soft assignments represent probability distributions over the set of clusters, which can retain uncertainty and act as a regularization to reduce overconfidence and improve robustness. 
To further avoid cluster imbalance, representation collapse, and degenerate solutions,  
we formulate the construction of pseudo-labels $\boldsymbol{\Psi}$ as an {\em entropy-regularized optimal transport problem}~\cite{asano2019self} that encourages uniform cluster utilization and maintains high entropy in the cluster label distribution:
\begin{equation}\label{eq:sinkhorn-objective}
\min_{\boldsymbol{\Psi} \in \mathbb{U}(\mathbf{a}, \mathbf{b})} \left\langle \boldsymbol{\Psi}, -\boldsymbol{\Psi}^\prime \right\rangle + \varepsilon \sum_{i,j} \boldsymbol{\Psi}_{i,j} \log \boldsymbol{\Psi}_{i,j},
\end{equation}
where $\mathbb{U}(\mathbf{a}, \mathbf{b})$ denotes the transportation polytope:
\begin{equation*}
\mathbb{U}(\mathbf{a}, \mathbf{b}) = \left\{ \boldsymbol{\Psi} \in \mathbb{R}^{|\V_u| \times K}_{\ge 0} \ \big| \ \boldsymbol{\Psi} \mathbf{1}_K = \mathbf{a}, \ \boldsymbol{\Psi}^\top \mathbf{1}_n = \mathbf{b} \right\},
\end{equation*}
with uniform marginals $\mathbf{a} = \frac{1}{n} \mathbf{1}_n$ and $\mathbf{b} = \frac{1}{K} \mathbf{1}_K$, and $\varepsilon > 0$ being the entropy regularization parameter. 
As proved in the literature~\cite{cuturi2013sinkhorn}, this problem can be solved by applying a fast {\em Sinkhorn-Knopp normalization} (SKN)~\cite{knight2008sinkhorn} over $\boldsymbol{\Psi}^\prime$ with $T$ iterations.
For the interest of space, we refer interested readers to Appendix~\ref{sec:SKN} in the supplementary materials~\cite{supplementary2025} for the algorithmic details.

%% file: tex/experiments.tex
\section{Experiments}

In this section, we experimentally evaluate \algo{} using classic GNN backbones against strong baselines for semi-supervised node classification. All experiments are conducted on a Linux machine equipped with an NVIDIA A100 GPU (80 GB memory), AMD EPYC 9754 GPUs, and 1.5 TB RAM.
Our source code is publicly available at \url{https://anonymous.4open.science/r/NCGC-0F52 }

\begin{table}[!t]
\centering
\renewcommand{\arraystretch}{0.8}
\caption{Statistics of Datasets.}\label{tbl:exp-data}
\vspace{-3mm}
\begin{tabular}{l|r|r|r|c}
	\hline
	{\bf Dataset} & \multicolumn{1}{c|}{\bf \#Nodes } & \multicolumn{1}{c|}{\bf \#Edges } & \multicolumn{1}{c|}{\bf \#Attributes } & \multicolumn{1}{c}{\bf \#Classes}  \\
	\hline
    {\em Cora}~\cite{SenNBGGE08coracitesser} &  2,708&  5,278&  1433&    7\\
    {\em CiteSeer}~\cite{SenNBGGE08coracitesser} & 3,327 & 4,522& 3,703 & 6   \\
    {\em PubMed}~\cite{SenNBGGE08coracitesser} &  19,717&  44,324&  500&    3\\
    {\em Computer}~\cite{shchur2018pitfalls}&  13,752&  245,861&  767&    10\\
    {\em  Photo}~\cite{shchur2018pitfalls}&  7,650&  119,081&  745&    8\\
    {\em CS}~\cite{shchur2018pitfalls}&  18,333&  81,894&  6,805&    15\\
    {\em  Physics}~\cite{shchur2018pitfalls}&  34,493&  247,962&  8,415&    5\\
    \hline
\end{tabular}%
\vspace{-1ex}
\end{table}

\subsection{Datasets}
We evaluate our approach on seven widely used benchmark datasets for semi-supervised node classification. These datasets include the citation networks {\em Cora}, {\em Citeseer}, and {\em PubMed}; Amazon co-purchase networks {\em Computer} and {\em Photo}; and co-authorship graphs {\em CS} and {\em Physics} extracted from the Microsoft Academic Graph~\cite{shchur2018pitfalls}.

\stitle{Citation Networks} Cora, Citeseer, and PubMed are three widely used citation graphs, where nodes represent documents and edges indicate citation relationships. For citation networks, we follow a similar semi-supervised setting in \cite{kipf2017semisupervised}, where each class randomly selects 20 labeled nodes for training, along with 500 nodes for validation and 1,000 for testing.

\stitle{Co-purchase Networks} Computer and Photo are product co-purchase graphs from Amazon, where nodes correspond to goods and edges indicate frequently co-bought items. As suggested by \citet{shchur2018pitfalls}, we adopt the class-balanced split protocol and randomly select 20 nodes per class for training and 30 nodes per class for validation, with all remaining nodes used for testing.

\stitle{Co-authorship Networks} CS and Physics are co-authorship graphs derived from the Microsoft Academic Graph, where nodes represent authors and edges indicate co-authorship. We use the same splitting strategy as for the Amazon datasets, with all splits generated independently for each run to ensure robust evaluation.

\input{figs/main-table}

\subsection{Baselines}
We compare our proposed method with a broad range of strong baselines for semi-supervised node classification. These baselines span across classic GNN architectures, deep or decoupled GNN variants, augmentation-based methods, and pseudo-labeling approaches:

\begin{itemize}[leftmargin=*]
    \item \textbf{Classic GNNs.} We include classical GNN models such as \texttt{GCN}~\cite{kipf2017semisupervised}, \texttt{GAT}~\cite{veličković2018graph}, \texttt{GraphSAGE}~\cite{hamilton2017inductive}, and \texttt{SGC}~\cite{wu2019simplifying}, along with their fully-finetuned variants (\texttt{GCN*}, \texttt{GAT*}, \texttt{GraphSAGE*}) as re-implemented in~\cite{luoclassic} under consistent evaluation protocols.

    \item \textbf{Deep GNNs.} To overcome limitations of shallow architectures, we consider \texttt{GCNII}~\cite{chen2020simple}, \texttt{APPNP}~\cite{gasteiger2018predict}, \texttt{DAGNN}~\cite{liu2020towards}, \texttt{JKNet}~\cite{xu2018representation}, and \texttt{GPR-GNN}~\cite{Chien21adaptive}, which adopt deeper or more flexible message passing mechanisms for improved representation learning.

    \item \textbf{Augmentation-based Methods.} Several methods leverage data augmentation or consistency regularization, including \texttt{GRAND}~\cite{feng2020graph}, \texttt{GraphMix}~\cite{verma2021graphmix}, \texttt{NodeAug}~\cite{wang2020nodeaug}, \texttt{GAM}~\cite{stretcu2019graph}, and \texttt{Violin}~\cite{xie2023violin}. We also include \texttt{MGCN}~\cite{yang2024mixed} and \texttt{AutoGDA}~\cite{zhao2022autogda}, which incorporate contrastive learning and automated data augmentation search strategies.

    \item \textbf{Pseudo-labeling Methods.} We include recent methods that utilize pseudo-labeling or confidence-driven training strategies, including \texttt{Meta-PN}~\cite{ding2022meta}, \texttt{M3S}~\cite{sun2020multi}, \texttt{DND-Net}~\cite{ding2024divide}, \texttt{NormProp}~\cite{zhang2025normalize}, and \texttt{PTA}~\cite{dong2021equivalence}. These methods aim to leverage unlabeled nodes by generating confident pseudo-labels or propagating soft labels in a principled manner.
\end{itemize}
All baseline results are carefully reproduced under our experimental setting. For data sets already tested in baselines, we follow the parameter settings provided by the respective authors, while other data sets are thoroughly fine-tuned. Due to space constraints, we defer detailed settings for hyperparameters in \algo{} to Appendix~\ref{sec:hyperparam-set} in the supplementary material~\cite{supplementary2025}. 

\input{figs/ablation}

\subsection{Semi-Supervised Classification Performance}
As shown in Table~\ref{results}, \algo{} consistently outperforms all baseline models across all datasets in terms of classification accuracy. Notably, \algo{} (\texttt{GCNII}) delivers substantial performance gains, achieving peak accuracies of 86.40\% on Cora and 82.03\% on Pubmed. Moreover, \algo{} (\texttt{GCN}) also demonstrates competitive performance on Citeseer and Coauthor-CS, surpassing state-of-the-art pseudo-labeling and data augmentation methods by margins exceeding 1.54\% and 1.52\%, respectively. The clear performance improvements further support the superiority of our self-supervised clustering mechanism compared to conventional label propagation and augmentation strategies. Typically, existing approaches often rely on random or heuristic neighbor sampling that fails to adapt to the target node’s local structural context. In contrast, \algo{} unifies spectral clustering with GNNs via a soft orthogonal constraint, enabling cluster-centered assignment optimization that explicitly aligns representation learning with semantically meaningful node groups. This unified cluster--MPNN strategy introduces richer and more reliable unsupervised signals, allowing the model to exploit structural regularities while suppressing noise from irrelevant neighbors, thereby yielding more robust predictions across diverse graph domains.

For \algo{} (\texttt{GAT}), the strong results further highlight \algo{}'s ability to promote generalization under limited supervision and its versatility across medium-size graph data sets. Intuitively, nodes in large graphs often reside in diverse structural contexts with neighbors of varying quality,  and \algo{}'s spectral clustering-based approach can better distinguish node features at different community levels, leading to more accurate and robust node representations.

\subsection{Ablation Study}
To validate the effectiveness of our SOGNs in \S~\ref{sec:model-layers} and self-supervised clustering module in \S~\ref{sec:SSC}, we conduct ablation studies on our \algo{} framework with four GNN backbones, i.e., \texttt{GCN}, \texttt{GAT}, \texttt{GraphSAGE}, and \texttt{GCNII}. Specifically, we investigate the impact of four key components: (i) SOC, (ii) deep embedded clustering loss $\mathcal{L}_{\texttt{KL}}$, (iii) pseudo-labeling loss $\mathcal{L}_{\text{PL}}$, and (iv) SKN. 
To study the effect of SOC, we set $\beta=0$ in Eq.~\eqref{eq:update-H-new}, and thus, SOGNs degrade as standard GCN, GAT, SAGE, or GCNII.
For $\mathcal{L}_{\texttt{KL}}$, we drop the clustering module and exclude the clustering loss from the framework.
For the $\mathcal{L}_{\text{PL}}$, we completely remove all pseudo-label loss modules. As for SKN, we disable this normalization process and directly employ the raw soft clustering assignments for computing the pseudo-labeling loss. The overall results are summarized in Table~\ref{tbl:ablation}.

Generally, we can observe that all ablation variants with part of components removed witness a clear drop of performance comparing with \algo{} model. In particular, \algo{} without pseudo-labeling loss commonly decrease badly among all datasets, which shows that pseudo-labeling loss significantly improves model accuracy. 
When removing SKN, the accuracy of each backbone variant decrease significantly, which indicates that applying over-confident and imbalanced clustering assignments as pseudo labels will lead to sub-optimal representation learning and harm model's performance. 
Moreover, the ablation results of removing the SOC and clustering loss clearly demonstrate the effectiveness of our proposed SOGNs and self-supervised clustering module, as dropping any part will harm model's performance. 

\input{figs/beta}

\input{figs/temperature}

\subsection{Parameter Analysis}

Firstly, we investigate the influence of the soft orthogonal constraint parameter $\beta$ in SOGNs on \algo{} (\texttt{GCN}) and \algo{} (\texttt{GAT}). 
Figures~\ref{fig:beta-GCN} and~\ref{fig:beta-GAT} present the node classification performance when varying $\beta$ from $0.001$ to $0.010$. 
For most datasets, increasing $\beta$ from a small value initially leads to improved classification accuracy, as the orthogonality constraint effectively encourages the learned cluster assignments to be more decorrelated and informative. 
The performance generally peaks within a moderate range (e.g., $\beta \approx 0.003$--$0.008$ for GCN and $\beta \approx 0.003$--$0.005$ for GAT), beyond which accuracy either saturates or slightly decreases. 
This degradation is likely due to the overly strong orthogonality constraint dominating the optimization, thereby limiting the model’s flexibility in fitting the label structure. 
These results suggest that setting $\beta$ within the above moderate range achieves a good balance between structural regularization and discriminative learning.

We further study the impact of the entropy regularization parameter $\varepsilon$ (Eq.~\eqref{eq:sinkhorn-objective}) in SKN. 
Figures~\ref{fig:temperature-GCN} and ~\ref{fig:temperature-GAT} show the node classification performance achieved by \algo{} (\texttt{GCN}) and \algo{} (\texttt{GAT}) on all seven datasets when varying $\varepsilon$ in the range from $0.01$ to $1.0$.
For these figures, we can observe that, as increasing $\varepsilon$ from a small value (e.g., 0.01), the classification accuracy improves rapidly and reaches an optimum around $\varepsilon = 0.03$ or $0.04$, beyond which the performance tends to degrade due to the target distributions becoming overly uniform. 
These empirical results suggest that running SKN with $\varepsilon$ in the range of $0.03$ to $0.04$ is favorable.

%% file: figs/main-table.tex
\begin{table*}[!t]
\centering
\begin{small}
\caption{Semi-supervised classification performance (\% test accuracy). We conduct {5} or {3} trials and report the mean accuracy and standard deviation over the trials. Data with \textsuperscript{†} are taken from the original paper. (best bolded and best baseline underlined)}
\label{results}
\vspace{-2ex}
\addtolength{\tabcolsep}{-0.2em}
\begin{tabular}{@{}clccccccc@{}}
\toprule
& \textbf{Model} & {\em Cora} & {\em Citeseer} & {\em PubMed} & {\em Computer}& {\em Photo }& {\em CS }& {\em Physics }\\ \midrule 
\multirow{7}{*}{\rotatebox[origin=c]{90}{Base GNNs}} & \texttt{GCN}~\cite{kipf2017semisupervised}  &  81.00 ± 0.29&   69.97 ± 0.65&  79.83 ± 1.29&   81.04 ± 0.65&   89.92 ± 0.42&   89.36 ± 0.80&   92.63 ± 0.67\\
& \texttt{GCN}$^\ast$ ~\cite{luoclassic}  &  84.60 ± 0.66&   72.42 ± 0.50&  80.52 ± 1.16&   81.11 ± 1.60&   90.34 ± 0.84&   {92.77 ± 0.10}&   92.83 ± 0.20\\
& \texttt{GAT}~\cite{veličković2018graph}  &  83.67 ± 0.96&   68.80 ± 0.85&  78.80 ± 1.11&   77.44 ± 0.84&   89.97 ± 0.04&   87.36 ± 0.56&   92.16 ± 0.60\\
& \texttt{GAT}$^\ast$ ~\cite{luoclassic}  &  83.82 ± 0.85&   70.94 ± 1.25&  79.06 ± 2.08&   81.36 ± 1.43&   91.43 ± 0.15&   92.42 ± 0.10&   89.37 ± 1.72\\
& \texttt{GraphSAGE}~\cite{hamilton2017inductive}  &  80.47 ± 0.55&   68.90 ± 1.11&  79.47 ± 0.31&   76.66 ± 0.90&   90.20 ± 0.53&   90.37 ± 0.61&   91.94 ± 0.86\\
& \texttt{GraphSAGE}$^\ast$ ~\cite{luoclassic}  &  83.46 ± 0.72&   70.96 ± 1.05&  78.76 ± 0.61&  \underline{82.10 ± 1.39} &   91.15 ± 0.55&   92.72 ± 0.09& 92.70 ± 0.46   \\
& \texttt{SGC}~\cite{wu2019simplifying}  &  80.00 ± 1.07&   65.74 ± 0.62&  76.23 ± 0.05&   72.84 ± 1.48&   84.51 ± 0.40&   87.67 ± 0.14&   91.32 ± 0.62\\
\midrule
\multirow{5}{*}{\rotatebox[origin=c]{90}{Deep GNNs}} & \texttt{GCNII}~\cite{chen2020simple}  &  77.79 ± 1.97&   65.16 ± 1.26&  76.18 ± 1.78&   74.72 ± 5.03&   87.78 ± 1.48&   89.87 ± 1.28&   92.75 ± 0.09\\
& \texttt{APPNP}~\cite{gasteiger2018predict}  &  80.36 ± 3.01&   68.14 ± 0.88&  77.29 ± 0.82&   73.11 ± 0.25&   86.98 ± 0.20&   90.53 ± 0.95&   93.53 ± 0.13\\
& \texttt{DAGNN}~\cite{liu2020towards}  &  80.38 ± 1.73&   66.90 ± 0.81&  77.50 ± 1.31&   80.81 ± 0.93&   90.78 ± 0.22&   92.70 ± 0.18&   93.39 ± 0.01\\
& \texttt{JKNet}~\cite{xu2018representation}  &  80.63 ± 0.17&   67.34 ± 0.76&  75.22 ± 0.84&   80.45 ± 0.53&   89.76 ± 0.12&   90.26 ± 0.05&   92.80 ± 0.25\\
& \texttt{GPR-GNN}~\cite{Chien21adaptive}&  81.51 ± 1.27&   67.21 ± 0.50&  78.30 ± 1.61&   79.52 ± 2.00&   90.81 ± 0.18&   \underline{92.79 ± 0.10}&   93.57 ± 0.11\\
\midrule
\multirow{5}{*}{\rotatebox[origin=c]{90}{Augmentation}} & \texttt{GRAND}~\cite{feng2020graph}  & 81.55 ± 0.27 & 70.92 ± 0.84  & 75.41 ± 0.51 &   69.76 ± 0.52&   82.75 ± 0.38&   90.80 ± 0.07&   92.64 ± 0.12\\
& \texttt{GraphMix}~\cite{verma2021graphmix} &  79.56 ± 0.86&   68.53 ± 0.45&  78.47 ± 1.34&   71.54 ± 0.45&   88.45 ± 0.76&   90.02 ± 0.58&   92.45 ± 0.84\\
& \texttt{GAM}~\cite{stretcu2019graph}  &  75.66 ± 0.15&   65.15 ± 0.04&  78.25 ± 0.25&   81.64 ± 0.36&   91.23 ± 0.32&   92.16 ± 0.84&   93.31 ± 1.51\\
& \texttt{Violin}\textsuperscript{\dag}~\cite{xie2023violin}  & 84.49 ± 0.66 & \underline{74.26 ± 0.40} & 81.23 ± 0.42 &  - & -  &  - &  - \\
& \texttt{MGCN}~\cite{yang2024mixed} &  81.23 ± 0.43&   67.17±0.95 &  76.34 ± 0.73&   54.77 ± 2.34&   90.56 ± 0.62&   90.42 ± 0.35&   OOM\\
\midrule
\multirow{5}{*}{\rotatebox[origin=c]{90}{Pseudo-label}} & \texttt{Meta-PN}~\cite{ding2022meta}  &  82.33 ± 0.79&   66.53 ± 1.13&  81.90 ± 0.45&   72.46 ± 7.32&   79.32 ± 12.71&   89.08 ± 0.34&   92.06 ± 1.23\\
& \texttt{M3S}~\cite{sun2020multi}  &  77.30 ± 0.46&   63.98 ± 1.11&  75.58 ± 1.15&   75.92 ± 1.07&   87.09 ± 0.52&   91.00 ± 0.20&   89.45 ± 0.95\\
& \texttt{DND-Net}~\cite{ding2024divide}  &  84.98 ± 0.56&  - &  78.24 ± 0.98&   - &   -&   76.07 ± 6.69&   OOM\\
& \texttt{NormProp}~\cite{zhang2025normalize} &  \underline{86.07 ± 0.00} &   72.83 ± 0.42&  \underline{81.53 ± 0.01}&   72.98 ± 4.64&   \underline{91.93 ± 0.44}&   90.79 ± 0.83&   \underline{94.28 ± 0.00}\\
& \texttt{PTA}~\cite{dong2021equivalence}  &  82.03 ± 1.05&   67.70 ± 0.73&  80.87 ± 0.62&   55.41 ± 2.79&   74.17 ± 1.57&   91.65 ± 0.44&   94.24 ± 0.31\\
\midrule
& \algo{} (\texttt{GCN}) &  {86.16 ± 0.81}&   \textbf{75.80 ± 0.29}&  81.68 ± 0.30&   83.23 ± 0.39&   92.34 ± 0.09&   \textbf{93.68 ± 0.24}&   \textbf{94.41 ± 0.29}\\
& \algo{} (\texttt{GAT}) &  84.72 ± 0.93&   {74.46 ± 1.11} & 79.62 ± 0.43&   \textbf{84.01 ± 0.94}&   \textbf{92.74 ± 0.38}&   93.09 ± 0.05&   93.88 ± 0.35\\
& \algo{} (\texttt{GraphSAGE}) &  84.84 ± 0.90&   74.24 ± 0.36&  79.84 ± 0.42&   {83.00 ± 1.42}&   {92.39 ± 0.03}&   {93.48 ± 0.18}&   93.89 ± 0.15\\
& \algo{} (\texttt{GCNII}) &  \textbf{86.40 ± 0.20}&   74.40 ± 0.26&  \textbf{82.03 ± 0.32}&  81.91 ± 0.09 &   91.02 ± 0.12&   91.35 ± 0.30&    94.14 ± 0.28\\
\bottomrule
\end{tabular}
\end{small}
\end{table*}

%% file: figs/ablation.tex
\begin{table*}[!t]
\centering
\begin{small}
\caption{Ablation study of \algo{} with various GNN backbones. }\label{tbl:ablation}
\vspace{-2ex}
\renewcommand{\arraystretch}{1.10}
\addtolength{\tabcolsep}{-0.2em}
\begin{tabular}{@{}lccccccc@{}}
\toprule
 \textbf{Model} & {\em Cora} & {\em Citeseer} & {\em PubMed} & {\em Computer} & {\em Photo} & {\em CS} & {\em Physics}\\ \midrule 
 w/o SOC&  \underline{85.72 ± 0.22}
&   75.34 ± 0.25&  80.60 ± 0.77
&   \underline{82.71 ± 0.67}&   90.91 ± 0.35&   \underline{92.81 ± 0.38}&   \underline{94.23 ± 0.35}\\
 w/o $\mathcal{L}_{\text{KL}}$ &  85.06 ± 0.59&   \underline{75.76 ± 0.32}&  80.82 ± 0.43&   80.65 ± 2.65&   \underline{91.00 ± 0.26}&   92.19 ± 1.32
&   90.24 ± 6.31\\
 w/o $\mathcal{L}_{\text{PL}}$&  83.30 ± 1.15
&  72.62 ± 0.60

&  81.20 ± 0.29
&   80.98 ± 2.22
&   90.97 ± 0.26&   33.30 ± 51.20&   91.90 ± 1.72\\
 w/o SKN& 83.78 ± 0.80& 72.98 ± 0.45& \underline{81.24 ± 0.49}& 81.26 ± 1.53& 91.00 ± 0.67& 92.03 ± 0.54&92.00 ± 0.30\\ \hline
 \algo{} (\texttt{GCN}) &  \textbf{86.16 ± 0.81} &   \textbf{75.80 ± 0.29}&  \textbf{81.68 ± 0.30}&   \textbf{83.23 ± 0.39}&   \textbf{92.34 ± 0.09}&   \textbf{93.68 ± 0.24}&   \textbf{94.41 ± 0.29}\\ \hline \hline
 w/o SOC&  \underline{83.76 ± 0.40}&   71.82 ± 0.48&  77.86 ± 1.11&   \underline{82.97 ± 1.25}&   \underline{91.84 ± 0.66}&   92.19 ± 0.20&   92.21 ± 0.33\\
 w/o $\mathcal{L}_{\text{KL}}$ &  83.58 ± 0.99&   \underline{74.18 ± 1.24}&  \underline{79.54 ± 0.62}&   82.51 ± 0.90
&   91.30 ± 1.32
&   \underline{92.88 ± 0.43}&   \underline{93.71 ± 0.44}\\
 w/o $\mathcal{L}_{\text{PL}}$&  83.02 ± 1.31&   71.22 ± 0.60&  78.52 ± 0.69&   81.62 ± 0.58&   91.52 ± 0.57&   92.15 ± 0.11&   92.65 ± 0.30\\
 w/o SKN& 83.42 ± 0.56& 71.96 ± 0.92& 78.48 ± 0.68
& 82.17 ± 1.15& 91.39 ± 0.69& 92.27 ± 0.34&92.72 ± 0.39\\ \hline
 \algo{} (\texttt{GAT}) &  \textbf{84.72 ± 0.93} &   \textbf{74.46 ± 1.11}&  \textbf{81.73 ± 0.31}&   \textbf{84.01 ± 0.94} &   \textbf{92.74 ± 0.38} &   \textbf{93.09 ± 0.05}&   \textbf{93.88 ± 0.35}\\ \hline \hline
  w/o SOC&  81.70 ± 0.65&   70.76 ± 1.09&  77.28 ± 0.84&   81.61 ± 1.11&   91.59 ± 0.41&   \underline{92.49 ± 0.10}&   92.97 ± 0.53\\
 w/o $\mathcal{L}_{\text{KL}}$ &  \underline{84.14 ± 1.36}&   \underline{73.94 ± 0.62}
&  \underline{79.00 ± 0.95}&   \underline{82.47 ± 0.52}&   \underline{92.20 ± 0.86}&   63.35 ± 51.62&   \underline{93.61 ± 0.41}\\
 w/o $\mathcal{L}_{\text{PL}}$&  82.32 ± 0.99&   71.96 ± 0.36&  77.52 ± 0.93&   79.28 ± 1.87&   91.28 ± 0.39&   92.29 ± 0.64&   93.22 ± 0.50\\
 w/o SKN& 82.12 ± 1.06& 71.56 ± 0.38
& 77.28 ± 2.13& 81.67 ± 0.78
& 91.40 ± 0.59& 92.31 ± 0.59&93.31 ± 0.14\\ \hline
 \algo{} (\texttt{GraphSAGE}) &  \textbf{84.84 ± 0.90}&   \textbf{74.24 ± 0.36}&  \textbf{79.84 ± 0.42}&   \textbf{83.00 ± 1.42}&   \textbf{92.39 ± 0.03}&   \textbf{93.48 ± 0.18}&   \textbf{93.89 ± 0.15}\\ \hline \hline
 w/o SOC&  85.57 ± 0.21&   70.53 ± 0.78&  \underline{81.60 ± 0.17}&   79.36 ± 0.66&   90.23 ± 0.33&   90.16 ± 0.12&   93.34 ± 0.61\\
 w/o $\mathcal{L}_{\text{KL}}$ &  \underline{86.03 ± 0.42}&   \underline{73.97 ± 0.35}
&81.57 ± 0.40  &   78.85 ± 2.91&   87.31 ± 1.76&   \underline{90.28 ± 0.30}&   \underline{93.91 ± 0.22}\\
 w/o $\mathcal{L}_{\text{PL}}$&  85.70 ± 0.17&   70.93 ± 1.42&  80.77 ± 0.21&   79.88 ± 0.66&   \underline{90.46 ± 0.32}&   82.11 ± 3.87&   93.59 ± 0.04\\
 w/o SKN& 85.30 ± 0.35& 71.57 ± 1.00& 81.33 ± 0.23& \underline{80.07 ± 0.93}& 87.94 ± 1.69& 89.17 ± 0.88&93.56 ± 0.09\\ \hline
 \algo{} (\texttt{GCNII}) &  \textbf{86.40 ± 0.20}&   \textbf{74.40 ± 0.26}&  \textbf{82.03 ± 0.32}&   \textbf{81.91 ± 0.09}&   \textbf{91.02 ± 0.12}&   \textbf{91.35 ± 0.30}&   \textbf{94.14 ± 0.28}\\ \hline
\end{tabular}
\end{small}
\vspace{0ex}
\end{table*}

%% file: figs/beta.tex
\begin{figure*}[!t]
\centering
\begin{small}
\resizebox{\textwidth}{!}{%
\subfloat[Cora]{
    \begin{tikzpicture}[scale=1]
    \begin{axis}[
        height=\columnwidth/2.5,
        width=\columnwidth/2.45,
        ylabel={\em Accuracy (\%)},
        xlabel={$\beta$},
        xmin=1, xmax=10,
        ymin=84.5, ymax=86.5,
        xtick={1,2,3,4,5,6,7,8,9,10},
        x tick label style={rotate=90,anchor=east},
        xticklabels={0.001,0.002,0.003,0.004,0.005,0.006,0.007,0.008,0.009,0.010},
        ytick={84.5, 85, 85.5, 86, 86.5},
        yticklabel style = {font=\tiny},
        xticklabel style = {font=\tiny},
        every axis label/.style={font=\footnotesize},
        title style={font=\footnotesize},
        grid=major,
        grid style={dashed, gray!30}
    ]
    \addplot[line width=0.4mm, mark=o, color=blue!80!black]
        coordinates {
            (1, 85.38)
            (2, 86.02)
            (3, 86.08)
            (4, 85.34)
            (5, 85.24)
            (6, 85.60)
            (7, 85.20)
            (8, 85.78)
            (9, 85.74)
            (10, 85.36)
        };
    \end{axis}
    \end{tikzpicture}
}
\subfloat[CiteSeer]{
    \begin{tikzpicture}[scale=1]
    \begin{axis}[
        height=\columnwidth/2.5,
        width=\columnwidth/2.45,
        xlabel={$\beta$},
        xmin=1, xmax=10,
        ymin=75.1, ymax=75.9,
        xtick={1,2,3,4,5,6,7,8,9,10},
        x tick label style={rotate=90,anchor=east},
        xticklabels={0.001,0.002,0.003,0.004,0.005,0.006,0.007,0.008,0.009,0.010},
        yticklabel style = {font=\tiny},
        xticklabel style = {font=\tiny},
        every axis label/.style={font=\footnotesize},
        title style={font=\footnotesize},
        grid=major,
        grid style={dashed, gray!30}
    ]
    \addplot[line width=0.4mm, mark=o, color=blue!80!black]
        coordinates {
            (1, 75.28)
            (2, 75.20)
            (3, 75.30)
            (4, 75.24)
            (5, 75.30)
            (6, 75.16)
            (7, 75.36)
            (8, 75.80)
            (9, 75.58)
            (10, 75.50)
        };
    \end{axis}
    \end{tikzpicture}
}
\subfloat[PubMed]{
    \begin{tikzpicture}[scale=1]
    \begin{axis}[
        height=\columnwidth/2.5,
        width=\columnwidth/2.45,
        xlabel={$\beta$},
        xmin=1, xmax=10,
        ymin=80, ymax=82,
        xtick={1,2,3,4,5,6,7,8,9,10},
        x tick label style={rotate=90,anchor=east},
        xticklabels={0.001,0.002,0.003,0.004,0.005,0.006,0.007,0.008,0.009,0.010},
        yticklabel style = {font=\tiny},
        xticklabel style = {font=\tiny},
        every axis label/.style={font=\footnotesize},
        title style={font=\footnotesize},
        grid=major,
        grid style={dashed, gray!30}
    ]
    \addplot[line width=0.4mm, mark=o, color=blue!80!black]
        coordinates {
            (1, 80.60)
            (2, 80.48)
            (3, 81.20)
            (4, 81.26)
            (5, 81.30)
            (6, 81.34)
            (7, 81.42)
            (8, 81.68)
            (9, 81.42)
            (10,81.50)
        };
    \end{axis}
    \end{tikzpicture}
}
\subfloat[Computer]{
    \begin{tikzpicture}[scale=1]
    \begin{axis}[
        height=\columnwidth/2.5,
        width=\columnwidth/2.45,
        ylabel={\em Accuracy (\%)},
        xlabel={$\beta$},
        xmin=1, xmax=10,
        ymin=79, ymax=84,
        xtick={1,2,3,4,5,6,7,8,9,10},
        xticklabels={0.001,0.002,0.003,0.004,0.005,0.006,0.007,0.008,0.009,0.010},
        xticklabel style={font=\tiny,rotate=90,anchor=east},
        yticklabel style={font=\tiny},
        every axis label/.style={font=\footnotesize},
        title style={font=\footnotesize},
        grid=major,
        grid style={dashed, gray!30}
    ]
    \addplot[line width=0.4mm, mark=o, color=blue!80!black]
        coordinates {
            (1, 81.51)
            (2, 82.34)
            (3, 81.20)
            (4, 83.23)
            (5, 82.92)
            (6, 81.88)
            (7, 81.38)
            (8, 81.25)
            (9, 79.39)
            (10, 81.22)
        };
    \end{axis}
    \end{tikzpicture}
}
\subfloat[Photo]{
    \begin{tikzpicture}[scale=1]
    \begin{axis}[
        height=\columnwidth/2.5,
        width=\columnwidth/2.45,
        xlabel={$\beta$},
        xmin=1, xmax=10,
        ymin=91.2, ymax=92.5,
        xtick={1,2,3,4,5,6,7,8,9,10},
        xticklabels={0.001,0.002,0.003,0.004,0.005,0.006,0.007,0.008,0.009,0.010},
        xticklabel style={font=\tiny,rotate=90,anchor=east},
        yticklabel style={font=\tiny},
        every axis label/.style={font=\footnotesize},
        title style={font=\footnotesize},
        grid=major,
        grid style={dashed, gray!30}
    ]
    \addplot[line width=0.4mm, mark=o, color=blue!80!black]
        coordinates {
            (1, 91.32)
            (2, 92.03)
            (3, 91.98)
            (4, 91.89)
            (5, 91.94)
            (6, 91.47)
            (7, 92.34)
            (8, 91.65)
            (9, 91.57)
            (10, 91.94)
        };
    \end{axis}
    \end{tikzpicture}
}
\subfloat[CS]{
    \begin{tikzpicture}[scale=1]
    \begin{axis}[
        height=\columnwidth/2.5,
        width=\columnwidth/2.45,
        xlabel={$\beta$},
        xmin=1, xmax=10,
        ymin=92.0, ymax=94.0,
        xtick={1,2,3,4,5,6,7,8,9,10},
        x tick label style={rotate=90,anchor=east},
        xticklabels={0.001,0.002,0.003,0.004,0.005,0.006,0.007,0.008,0.009,0.010},
        yticklabel style = {font=\tiny},
        xticklabel style = {font=\tiny},
        every axis label/.style={font=\footnotesize},
        title style={font=\footnotesize},
        grid=major,
        grid style={dashed, gray!30}
    ]
    \addplot[line width=0.4mm, mark=o, color=blue!80!black]
        coordinates {
            (1, 92.97)
            (2, 92.55)
            (3, 93.56)
            (5, 93.26)
            (6, 92.71)
            (7, 93.19)
            (8, 93.38)
            (9, 92.96)
            (10, 93.19)
        };
    \end{axis}
    \end{tikzpicture}
}
\subfloat[Physics]{
    \begin{tikzpicture}[scale=1]
    \begin{axis}[
        height=\columnwidth/2.5,
        width=\columnwidth/2.45,
        xlabel={$\beta$},
        xmin=1, xmax=10,
        ymin=92.8, ymax=95,
        xtick={1,2,3,4,5,6,7,8,9,10},
        x tick label style={rotate=90,anchor=east},
        xticklabels={0.001,0.002,0.003,0.004,0.005,0.006,0.007,0.008,0.009,0.010},
        yticklabel style = {font=\tiny},
        xticklabel style = {font=\tiny},
        every axis label/.style={font=\footnotesize},
        title style={font=\footnotesize},
        grid=major,
        grid style={dashed, gray!30}
    ]
    \addplot[line width=0.4mm, mark=o, color=blue!80!black]
        coordinates {
            (1, 93.53)
            (2, 93.63)
            (3, 93.95)
            (4, 93.92)
            (5, 94.10)
            (6, 93.84)
            (7, 93.99)
            (8, 94.41)
            (9, 92.95)
            (10, 93.57)
        };
    \end{axis}
    \end{tikzpicture}
}
}
\end{small}
\vspace{-2ex}
\caption{Varying $\beta$ in \algo{} (\texttt{GCN}).}
\label{fig:beta-GCN}
\vspace{-2ex}
\end{figure*}

\begin{figure*}[!t]
\centering
\begin{small}
\resizebox{\textwidth}{!}{%
\subfloat[Cora]{
    \begin{tikzpicture}[scale=1]
    \begin{axis}[
        height=\columnwidth/2.5,
        width=\columnwidth/2.45,
        ylabel={\em Accuracy (\%)},
        xlabel={$\beta$},
        xmin=1, xmax=10,
        ymin=83.5, ymax=84.5,
        xtick={1,2,3,4,5,6,7,8,9,10},
        x tick label style={rotate=90,anchor=east},
        xticklabels={0.001,0.002,0.003,0.004,0.005,0.006,0.007,0.008,0.009,0.010},
        ytick={83.5, 83.75, 84, 84.25, 84.5},
        yticklabel style = {font=\tiny},
        xticklabel style = {font=\tiny},
        every axis label/.style={font=\footnotesize},
        title style={font=\footnotesize},
        grid=major,
        grid style={dashed, gray!30}
    ]
    \addplot[line width=0.4mm, mark=o, color=blue!80!black]
        coordinates {
            (1, 84.04)
            (2, 84.00)
            (3, 84.38)
            (4, 84.26)
            (5, 83.82)
            (6, 84.26)
            (7, 84.22)
            (8, 84.08)
            (9, 83.68)
            (10, 83.78)
        };
    \end{axis}
    \end{tikzpicture}
}
\subfloat[CiteSeer]{
    \begin{tikzpicture}[scale=1]
    \begin{axis}[
        height=\columnwidth/2.5,
        width=\columnwidth/2.45,
        xlabel={$\beta$},
        xmin=1, xmax=10,
        ymin=73.5, ymax=75,
        xtick={1,2,3,4,5,6,7,8,9,10},
        x tick label style={rotate=90,anchor=east},
        xticklabels={0.001,0.002,0.003,0.004,0.005,0.006,0.007,0.008,0.009,0.010},
        yticklabel style = {font=\tiny},
        xticklabel style = {font=\tiny},
        every axis label/.style={font=\footnotesize},
        title style={font=\footnotesize},
        grid=major,
        grid style={dashed, gray!30}
    ]
    \addplot[line width=0.4mm, mark=o, color=blue!80!black]
        coordinates {
            (1, 74.26)
            (2, 74.44)
            (3, 74.38)
            (4, 74.14)
            (5, 74.42)
            (6, 73.60)
            (7, 73.88)
            (8, 74.12)
            (9, 74.02)
            (10, 73.88)
        };
    \end{axis}
    \end{tikzpicture}
}
\subfloat[PubMed]{
    \begin{tikzpicture}[scale=1]
    \begin{axis}[
        height=\columnwidth/2.5,
        width=\columnwidth/2.45,
        xlabel={$\beta$},
        xmin=1, xmax=10,
        ymin=77.5, ymax=80,
        xtick={1,2,3,4,5,6,7,8,9,10},
        x tick label style={rotate=90,anchor=east},
        xticklabels={0.001,0.002,0.003,0.004,0.005,0.006,0.007,0.008,0.009,0.010},
        yticklabel style = {font=\tiny},
        xticklabel style = {font=\tiny},
        every axis label/.style={font=\footnotesize},
        title style={font=\footnotesize},
        grid=major,
        grid style={dashed, gray!30}
    ]
    \addplot[line width=0.4mm, mark=o, color=blue!80!black]
        coordinates {
            (1, 78.88)
            (2, 78.24)
            (3, 78.74)
            (4, 78.56)
            (5, 79.20)
            (6, 79.32)
            (7, 79.34)
            (8, 79.62)
            (9, 78.96)
            (10, 77.96)
        };
    \end{axis}
    \end{tikzpicture}
}
\subfloat[Computer]{
    \begin{tikzpicture}[scale=1]
    \begin{axis}[
        height=\columnwidth/2.5,
        width=\columnwidth/2.45,
        ylabel={\em Accuracy (\%)},
        xlabel={$\beta$},
        xmin=1, xmax=10,
        ymin=82, ymax=85,
        xtick={1,2,3,4,5,6,7,8,9,10},
        xticklabels={0.001,0.002,0.003,0.004,0.005,0.006,0.007,0.008,0.009,0.010},
        xticklabel style={font=\tiny,rotate=90,anchor=east},
        yticklabel style={font=\tiny},
        every axis label/.style={font=\footnotesize},
        title style={font=\footnotesize},
        grid=major,
        grid style={dashed, gray!30}
    ]
    \addplot[line width=0.4mm, mark=o, color=blue!80!black]
        coordinates {
            (1, 82.99)
            (2, 83.20)
            (3, 83.93)
            (4, 83.56)
            (5, 84.01)
            (6, 83.95)
            (7, 83.26)
            (8, 83.26)
            (9, 83.88)
            (10, 83.91)
        };
    \end{axis}
    \end{tikzpicture}
}
\subfloat[Photo]{
    \begin{tikzpicture}[scale=1]
    \begin{axis}[
        height=\columnwidth/2.5,
        width=\columnwidth/2.45,
        xlabel={$\beta$},
        xmin=1, xmax=10,
        ymin=90.5, ymax=93,
        xtick={1,2,3,4,5,6,7,8,9,10},
        xticklabels={0.001,0.002,0.003,0.004,0.005,0.006,0.007,0.008,0.009,0.010},
        xticklabel style={font=\tiny,rotate=90,anchor=east},
        yticklabel style={font=\tiny},
        every axis label/.style={font=\footnotesize},
        title style={font=\footnotesize},
        grid=major,
        grid style={dashed, gray!30}
    ]
    \addplot[line width=0.4mm, mark=o, color=blue!80!black]
        coordinates {
            (1, 91.75)
            (2, 91.94)
            (3, 92.74)
            (4, 91.70)
            (5, 92.01)
            (6, 91.49)
            (7, 90.94)
            (8, 91.06)
            (9, 91.61)
            (10, 91.07)
        };
    \end{axis}
    \end{tikzpicture}
}
\subfloat[CS]{
    \begin{tikzpicture}[scale=1]
    \begin{axis}[
        height=\columnwidth/2.5,
        width=\columnwidth/2.45,
        xlabel={$\beta$},
        xmin=1, xmax=10,
        ymin=92.0, ymax=93.2,
        xtick={1,2,3,4,5,6,7,8,9,10},
        x tick label style={rotate=90,anchor=east},
        xticklabels={0.001,0.002,0.003,0.004,0.005,0.006,0.007,0.008,0.009,0.010},
        yticklabel style = {font=\tiny},
        xticklabel style = {font=\tiny},
        every axis label/.style={font=\footnotesize},
        title style={font=\footnotesize},
        grid=major,
        grid style={dashed, gray!30}
    ]
    \addplot[line width=0.4mm, mark=o, color=blue!80!black]
        coordinates {
            (1, 92.83)
            (2, 92.76)
            (3, 92.71)
            (4, 93.09)
            (5, 92.15)
            (6, 92.95)
            (7, 92.87)
            (8, 92.95)
            (9, 92.93)
            (10, 92.19)
        };
    \end{axis}
    \end{tikzpicture}
}
\subfloat[Physics]{
    \begin{tikzpicture}[scale=1]
    \begin{axis}[
        height=\columnwidth/2.5,
        width=\columnwidth/2.45,
        xlabel={$\beta$},
        xmin=1, xmax=10,
        ymin=93.5, ymax=94.0,
        xtick={1,2,3,4,5,6,7,8,9,10},
        x tick label style={rotate=90,anchor=east},
        xticklabels={0.001,0.002,0.003,0.004,0.005,0.006,0.007,0.008,0.009,0.010},
        yticklabel style = {font=\tiny},
        xticklabel style = {font=\tiny},
        every axis label/.style={font=\footnotesize},
        title style={font=\footnotesize},
        grid=major,
        grid style={dashed, gray!30}
    ]
    \addplot[line width=0.4mm, mark=o, color=blue!80!black]
        coordinates {
            (1, 93.89)
            (2, 93.85)
            (3, 93.81)
            (4, 93.89)
            (5, 93.85)
            (6, 93.83)
            (7, 93.82)
            (8, 93.61)
            (9, 93.59)
            (10, 93.56)
        };
    \end{axis}
    \end{tikzpicture}
}
}
\end{small}
\vspace{-2ex}
\caption{Varying $\beta$ in \algo{} (\texttt{GAT}).}
\label{fig:beta-GAT}
\vspace{-2ex}
\end{figure*}

%% file: figs/temperature.tex
\begin{figure*}[!t]
\centering
\begin{small}
\resizebox{\textwidth}{!}{%
\subfloat[Cora]{
    \begin{tikzpicture}[scale=1]
    \begin{axis}[
        height=\columnwidth/2.5,
        width=\columnwidth/2.45,
        ylabel={\em Accuracy (\%)},
        xlabel={$\varepsilon$},
        xmin=1, xmax=10,
        ymin=84.5, ymax=86.5,
        xtick={1,2,3,4,5,6,7,8,9,10},
        xticklabels={0.01,0.02,0.03,0.04,0.08,0.1,0.2,0.4,0.8,1},
        ytick={84.5,85,85.5,86,86.5},
        yticklabel style = {font=\tiny},
        xticklabel style = {font=\tiny,rotate=90,anchor=east},
        every axis label/.style={font=\footnotesize},
        title style={font=\footnotesize},
        grid=major,
        grid style={dashed, gray!30}
    ]
    \addplot[line width=0.4mm, mark=o, color=blue!80!black]
        coordinates {
            (1, 85.12)
            (2, 85.30)
            (3, 84.88)
            (4, 86.16)
            (5, 84.90)
            (6, 84.78)
            (7, 85.18)
            (8, 85.10)
            (9, 85.16)
            (10, 85.06)
        };
    \end{axis}
    \end{tikzpicture}
}
\subfloat[CiteSeer]{
    \begin{tikzpicture}[scale=1]
    \begin{axis}[
        height=\columnwidth/2.5,
        width=\columnwidth/2.45,
        xlabel={$\varepsilon$},
        xmin=1, xmax=10,
        ymin=0, ymax=85,
        xtick={1,2,3,4,5,6,7,8,9,10},
        x tick label style={rotate=90,anchor=east},
        xticklabels={0.01,0.02,0.03,0.04,0.08,0.1,0.2,0.4,0.8,1.0},
        yticklabel style = {font=\tiny},
        xticklabel style = {font=\tiny},
        every axis label/.style={font=\footnotesize},
        title style={font=\footnotesize},
        grid=major,
        grid style={dashed, gray!30}
    ]
    \addplot[line width=0.4mm, mark=o, color=blue!80!black]
        coordinates {
            (1, 7.20)
            (2, 75.53)
            (3, 75.40)
            (4, 74.93)
            (5, 73.27)
            (6, 73.20)
            (7, 72.73)
            (8, 72.23)
            (9, 72.10)
            (10, 72.17)
        };
    \end{axis}
    \end{tikzpicture}
}
\subfloat[PubMed]{
    \begin{tikzpicture}[scale=1]
    \begin{axis}[
        height=\columnwidth/2.5,
        width=\columnwidth/2.45,
        xlabel={$\varepsilon$},
        xmin=1, xmax=10,
        ymin=80, ymax=81.5,
        xtick={1,2,3,4,5,6,7,8,9,10},
        x tick label style={rotate=90,anchor=east},
        xticklabels={0.01,0.02,0.03,0.04,0.08,0.1,0.2,0.4,0.8,1.0},
        yticklabel style = {font=\tiny},
        xticklabel style = {font=\tiny},
        every axis label/.style={font=\footnotesize},
        title style={font=\footnotesize},
        grid=major,
        grid style={dashed, gray!30}
    ]
    \addplot[line width=0.4mm, mark=o, color=blue!80!black]
        coordinates {
            (1, 21.50)
            (2, 81.20)
            (3, 81.07)
            (4, 81.27)
            (5, 80.87)
            (6, 80.73)
            (7, 80.83)
            (8, 80.37)
            (9, 80.80)
            (10, 80.83)
        };
    \end{axis}
    \end{tikzpicture}
}
\subfloat[Computer]{
    \begin{tikzpicture}[scale=1]
    \begin{axis}[
        height=\columnwidth/2.5,
        width=\columnwidth/2.45,
        xlabel={$\varepsilon$},
        xmin=1, xmax=9,
        ymin=80, ymax=83,
        xtick={1,2,3,4,5,6,7,8,9,10},
        x tick label style={rotate=90,anchor=east},
        xticklabels={0.01,0.02,0.03,0.04,0.08,0.1,0.2,0.4,0.8,1.0},
        yticklabel style = {font=\tiny},
        xticklabel style = {font=\tiny},
        every axis label/.style={font=\footnotesize},
        title style={font=\footnotesize},
        grid=major,
        grid style={dashed, gray!30}
    ]
    \addplot[line width=0.4mm, mark=o, color=blue!80!black]
        plot coordinates {
            (1, 82.47) (2, 82.55) (3,82.30 )(4, 82.79) (5, 82.29) (6, 82.37) (7, 82.57) (8, 82.34) (9, 81.64) (10, 80.48)
        };
    \end{axis}
    \end{tikzpicture}
}
\subfloat[Photo]{
    \begin{tikzpicture}[scale=1]
    \begin{axis}[
        height=\columnwidth/2.5,
        width=\columnwidth/2.45,
        xlabel={$\varepsilon$},
        xmin=1, xmax=9,
        ymin=89, ymax=92,
        xtick={1,2,3,4,5,6,7,8,9},
        x tick label style={rotate=90,anchor=east},
        xticklabels={0.01,0.02,0.03,0.04,0.08,0.1,0.2,0.4,0.8,1.0},
        ytick={89, 90, 91, 92},
        yticklabel style = {font=\tiny},
        xticklabel style = {font=\tiny},
        every axis label/.style={font=\footnotesize},
        title style={font=\footnotesize},
        grid=major,
        grid style={dashed, gray!30}
    ]
    \addplot[line width=0.4mm, mark=o, color=blue!80!black]
        plot coordinates {
            (1, 90.53)
            (2, 90.49)
            (3, 91.26)
            (4, 91.39)
            (5, 90.09)
            (6, 89.61)
            (7, 90.58)
            (8, 89.83)
            (9, 90.38)
            (10, 90.77)
        };
    \end{axis}
    \end{tikzpicture}
}
\subfloat[CS]{
    \begin{tikzpicture}[scale=1]
    \begin{axis}[
        height=\columnwidth/2.5,
        width=\columnwidth/2.45,
        xlabel={$\varepsilon$},
        xmin=1, xmax=10,
        ymin=0, ymax=100,
        xtick={1,2,3,4,5,6,7,8,9,10},
        x tick label style={rotate=90,anchor=east},
        xticklabels={0.01,0.02,0.03,0.04,0.08,0.1,0.2,0.4,0.8,1.0},
        yticklabel style = {font=\tiny},
        xticklabel style = {font=\tiny},
        every axis label/.style={font=\footnotesize},
        title style={font=\footnotesize},
        grid=major,
        grid style={dashed, gray!30}
    ]
    \addplot[line width=0.4mm, mark=o, color=blue!80!black]
        coordinates {
            (1, 3.74)
            (2, 3.74)
            (3, 69.13)
            (4, 93.68)
            (5, 92.58)
            (6, 92.28)
            (7, 89.30)
            (8, 88.06)
            (9, 86.87)
            (10, 85.21)
        };
    \end{axis}
    \end{tikzpicture}
}
\subfloat[Physics]{
    \begin{tikzpicture}[scale=1]
    \begin{axis}[
        height=\columnwidth/2.5,
        width=\columnwidth/2.45,
        xlabel={$\varepsilon$},
        xmin=1, xmax=10,
        ymin=15, ymax=100,
        xtick={1,2,3,4,5,6,7,8,9,10},
        x tick label style={rotate=90,anchor=east},
        xticklabels={0.01,0.02,0.03,0.04,0.08,0.1,0.2,0.4,0.8,1.0},
        yticklabel style = {font=\tiny},
        xticklabel style = {font=\tiny},
        every axis label/.style={font=\footnotesize},
        title style={font=\footnotesize},
        grid=major,
        grid style={dashed, gray!30}
    ]
    \addplot[line width=0.4mm, mark=o, color=blue!80!black]
        coordinates {
            (1, 16.65)
            (2, 79.48)
            (3, 93.85)
            (4, 94.41)
            (5, 93.64)
            (6, 93.43)
            (7, 88.30)
            (8, 85.95)
            (9, 89.56)
            (10, 87.67)
        };
    \end{axis}
    \end{tikzpicture}
}
}
\end{small}
\vspace{-2ex}
\caption{Varying entropy regularization parameter $\varepsilon$ in \algo{} (\texttt{GCN}).}
\label{fig:temperature-GCN}
\vspace{-2ex}
\end{figure*}

\begin{figure*}[!t]
\centering
\begin{small}
\resizebox{\textwidth}{!}{%
\subfloat[Cora]{
    \begin{tikzpicture}[scale=1]
    \begin{axis}[
        height=\columnwidth/2.5,
        width=\columnwidth/2.45,
        ylabel={\em Accuracy (\%)},
        xlabel={$\varepsilon$},
        xmin=1, xmax=10,
        ymin=11, ymax=90,
        xtick={1,2,3,4,5,6,7,8,9,10},
        x tick label style={rotate=90,anchor=east},
        xticklabels={0.01,0.02,0.03,0.04,0.08,0.1,0.2,0.4,0.8,1.0},
        yticklabel style = {font=\tiny},
        xticklabel style = {font=\tiny},
        every axis label/.style={font=\footnotesize},
        title style={font=\footnotesize},
        grid=major,
        grid style={dashed, gray!30}
    ]
    \addplot[line width=0.4mm, mark=o, color=blue!80!black]
        coordinates {
            (1, 11.40)
            (2, 84.40)
            (3, 84.72)
            (4, 83.88)
            (5, 83.78)
            (6, 83.72)
            (7, 83.12)
            (8, 82.66)
            (9, 82.44)
            (10, 81.90)
        };
    \end{axis}
    \end{tikzpicture}
}
\subfloat[CiteSeer]{
    \begin{tikzpicture}[scale=1]
    \begin{axis}[
        height=\columnwidth/2.5,
        width=\columnwidth/2.45,
        xlabel={$\varepsilon$},
        xmin=1, xmax=10,
        ymin=73.5, ymax=75,
        xtick={1,2,3,4,5,6,7,8,9,10},
        x tick label style={rotate=90,anchor=east},
        xticklabels={0.01,0.02,0.03,0.04,0.08,0.1,0.2,0.4,0.8,1.0},
        yticklabel style = {font=\tiny},
        xticklabel style = {font=\tiny},
        every axis label/.style={font=\footnotesize},
        title style={font=\footnotesize},
        grid=major,
        grid style={dashed, gray!30}
    ]
    \addplot[line width=0.4mm, mark=o, color=blue!80!black]
        coordinates {
            (1, 74.22)
            (2, 74.30)
            (3, 74.46)
            (4, 74.44)
            (5, 73.94)
            (6, 74.46)
            (7, 74.26)
            (8, 74.28)
            (9, 74.42)
            (10, 74.42)
        };
    \end{axis}
    \end{tikzpicture}
}
\subfloat[PubMed]{
    \begin{tikzpicture}[scale=1]
    \begin{axis}[
        height=\columnwidth/2.5,
        width=\columnwidth/2.45,
        xlabel={$\varepsilon$},
        xmin=1, xmax=10,
        ymin=77.5, ymax=80,
        xtick={1,2,3,4,5,6,7,8,9,10},
        x tick label style={rotate=90,anchor=east},
        xticklabels={0.01,0.02,0.03,0.04,0.08,0.1,0.2,0.4,0.8,1.0},
        yticklabel style = {font=\tiny},
        xticklabel style = {font=\tiny},
        every axis label/.style={font=\footnotesize},
        title style={font=\footnotesize},
        grid=major,
        grid style={dashed, gray!30}
    ]
    \addplot[line width=0.4mm, mark=o, color=blue!80!black]
        coordinates {
            (1, 79.48)
            (2, 79.30)
            (3, 77.86)
            (4, 79.50)
            (5, 79.34)
            (6, 79.32)
            (7, 79.34)
            (8, 79.36)
            (9, 79.36)
            (10, 79.40)
        };
    \end{axis}
    \end{tikzpicture}
}
\subfloat[Computer]{
    \begin{tikzpicture}[scale=1]
    \begin{axis}[
        height=\columnwidth/2.5,
        width=\columnwidth/2.45,
        xlabel={$\varepsilon$},
        xmin=1, xmax=9,
        ymin=80, ymax=83,
        xtick={1,2,3,4,5,6,7,8,9,10},
        x tick label style={rotate=90,anchor=east},
        xticklabels={0.01,0.02,0.03,0.04,0.08,0.1,0.2,0.4,0.8,1.0},
        yticklabel style = {font=\tiny},
        xticklabel style = {font=\tiny},
        every axis label/.style={font=\footnotesize},
        title style={font=\footnotesize},
        grid=major,
        grid style={dashed, gray!30}
    ]
    \addplot[line width=0.4mm, mark=o, color=blue!80!black]
        plot coordinates {
            (1, 82.47) (2, 82.55)(3,82.81) (4, 82.79) (5, 82.29) (6, 82.37) (7, 82.57) (8, 82.34) (9, 81.64) (10, 80.48)
        };
    \end{axis}
    \end{tikzpicture}
}
\subfloat[Photo]{
    \begin{tikzpicture}[scale=1]
    \begin{axis}[
        height=\columnwidth/2.5,
        width=\columnwidth/2.45,
        xlabel={$\varepsilon$},
        xmin=1, xmax=10,
        ymin=0, ymax=100,
        xtick={1,2,3,4,5,6,7,8,9,10},
        x tick label style={rotate=90,anchor=east},
        xticklabels={0.01,0.02,0.03,0.04,0.08,0.1,0.2,0.4,0.8,1.0},
        yticklabel style = {font=\tiny},
        xticklabel style = {font=\tiny},
        every axis label/.style={font=\footnotesize},
        title style={font=\footnotesize},
        grid=major,
        grid style={dashed, gray!30}
    ]
    \addplot[line width=0.4mm, mark=o, color=blue!80!black]
        coordinates {
            (1, 4.40)
            (2, 63.35)
            (3, 92.74)
            (4, 92.25)
            (5, 90.92)
            (6, 90.29)
            (7, 88.66)
            (8, 87.83)
            (9, 88.03)
            (10, 89.29)
        };
    \end{axis}
    \end{tikzpicture}
}
\subfloat[CS]{
    \begin{tikzpicture}[scale=1]
    \begin{axis}[
        height=\columnwidth/2.5,
        width=\columnwidth/2.45,
        xlabel={$\varepsilon$},
        xmin=1, xmax=10,
        ymin=3, ymax=100,
        xtick={1,2,3,4,5,6,7,8,9,10},
        x tick label style={rotate=90,anchor=east},
        xticklabels={0.01,0.02,0.03,0.04,0.08,0.1,0.2,0.4,0.8,1.0},
        yticklabel style = {font=\tiny},
        xticklabel style = {font=\tiny},
        every axis label/.style={font=\footnotesize},
        title style={font=\footnotesize},
        grid=major,
        grid style={dashed, gray!30}
    ]
    \addplot[line width=0.4mm, mark=o, color=blue!80!black]
        coordinates {
            (1, 3.74)
            (2, 3.74)
            (3, 92.93)
            (4, 93.07)
            (5, 92.41)
            (6, 92.55)
            (7, 91.72)
            (8, 91.16)
            (9, 91.54)
            (10, 90.87)
        };
    \end{axis}
    \end{tikzpicture}
}
\subfloat[Physics]{
    \begin{tikzpicture}[scale=1]
    \begin{axis}[
        height=\columnwidth/2.5,
        width=\columnwidth/2.45,
        xlabel={$\varepsilon$},
        xmin=1, xmax=10,
        ymin=93.70, ymax=93.95,
        xtick={1,2,3,4,5,6,7,8,9,10},
        x tick label style={rotate=90,anchor=east},
        xticklabels={0.01,0.02,0.03,0.04,0.08,0.1,0.2,0.4,0.8,1.0},
        yticklabel style = {font=\tiny},
        xticklabel style = {font=\tiny},
        every axis label/.style={font=\footnotesize},
        title style={font=\footnotesize},
        grid=major,
        grid style={dashed, gray!30}
    ]
    \addplot[line width=0.4mm, mark=o, color=blue!80!black]
        coordinates {
            (1, 93.76)
            (2, 93.85)
            (3, 93.88)
            (4, 93.79)
            (5, 93.83)
            (6, 93.75)
            (7, 93.87)
            (8, 93.86)
            (9, 93.76)
            (10, 93.76)
        };
    \end{axis}
    \end{tikzpicture}
}
}
\end{small}
\vspace{-2ex}
\caption{Varying entropy regularization parameter $\varepsilon$ in \algo{} (\texttt{GAT}).}
\label{fig:temperature-GAT}
\vspace{-2ex}
\end{figure*}

%% file: tex/appendix.tex
\section{Sinkhorn-Knopp Normalization}\label{sec:SKN}
The optimization objective in Eq.~\eqref{eq:sinkhorn-objective} can be efficiently solved using the Sinkhorn-Knopp algorithm~\cite{cuturi2013sinkhorn}, which alternates between row and column normalization. We first compute the unnormalized assignment matrix by exponentiating the scaled logits:
\begin{equation}\label{eq:init-sinkhorn}
\boldsymbol{\Psi}^{(0)} = \exp\left(\frac{\boldsymbol{\Psi}^\prime}{\varepsilon}\right)
\end{equation}
and then applying iterative normalization to \(\boldsymbol{\Psi}^{(0)}\) by alternately rescaling its rows and columns: 
\begin{equation}\label{eq:sinkhorn-iteration}
\boldsymbol{\Psi}^{(t+1)} = \mathrm{diag}\left(\frac{\mathbf{a}}{\boldsymbol{\Psi}^{(t)} \mathbf{1}_K}\right) \boldsymbol{\Psi}^{(t)} \mathrm{diag}\left(\frac{\mathbf{b}}{(\boldsymbol{\Psi}^{(t)})^\top \mathbf{1}_n}\right),
\end{equation}
until convergence. The resulting matrix $\boldsymbol{\Psi}$ lies in $\mathcal{U}(\mathbf{a}, \mathbf{b})$ and approximates a doubly stochastic matrix with balanced marginals and maximal entropy. In particular, the obtained $\boldsymbol{\Psi}$ has high marginal entropy due to the uniform constraint $\boldsymbol{\Psi}^\top \mathbf{1}_n = \mathbf{b}$, i.e., $H(\boldsymbol{\Psi}) \approx \log K$. Considering the mutual information between the node embeddings $\HM$ and soft cluster assignments $\boldsymbol{\Psi}$:
\begin{equation}\label{eq:mutual-info}
I(\HM; \boldsymbol{\Psi}) = H(\boldsymbol{\Psi}) - H(\boldsymbol{\Psi} \mid \HM),
\end{equation}
where $H(\boldsymbol{\Psi})$ denotes the marginal entropy and $H(\boldsymbol{\Psi} \mid \HM)$ the conditional entropy, it can be interpreted as being maximized by (i) enforcing high marginal entropy via the Sinkhorn constraint, and (ii) minimizing conditional entropy $H(\boldsymbol{\Psi} \mid \HM)$ by learning the model logits $\boldsymbol{\Psi}^\prime$ to predict cluster assignments more confidently. This mutual information maximization encourages each cluster to cover semantically meaningful regions of the embedding space while avoiding overconfident soft labels. 

\eat{
\renchi{why this can lead to balanced clusters/classifications? see \url{https://zhuanlan.zhihu.com/p/543052563}.}
\songbo{In general, we have already formulated objective into optimal transporation, and we manually add one equipartition constraint U(a,b). In other words, we have a prior belief that we assumpt the label distribution is uniform, and we just add this constraint. After adding this constraint, the optimal transportation problem is simplified and can be solved by sinkhorn. Shall I add some balanced cluster explanation to the sinkhorn part?}

\renchi{this is not theoretical analysis. needs math, formulas, lemmata or theorems.}

}

\section{Hyperparameter Settings}\label{sec:hyperparam-set}

\noindent
We provide the hyperparameter search strategy used for \algo{} from table~\ref{tab:hyperparams-GCN} to table~\ref{tab:hyperparams-GCNII}. We perform grid search over all key hyperparameters for each dataset. Specifically:

\begin{itemize}
    \item The \textbf{soft orthogonal constraint parameter} $\beta$ is searched in the range 0.001--0.010.
    \item The \textbf{entropy regularization parameter} $\epsilon$ is searched from 0.01 to 1.0.
    \item The \textbf{number of Sinkhorn iterations} $T$ is searched from 1 to 30.
    \item The \textbf{hidden dimension} is selected from \{64, 128, 256, 512\}.
    \item The \textbf{learning rate} is searched from \{0.001, 0.005, 0.01\}.
    \item The \textbf{dropout rate} is searched in the range 0.0--0.9.
    \item The \textbf{number of layers} is searched differently depending on model depth: for shallow architectures such as GCN, GAT, and GraphSAGE, we search from 1 to 6 layers; for deep architectures such as GCNII, we search from 2 to 64 layers.
\end{itemize}

These ranges are applied consistently across all datasets and models to identify the optimal hyperparameter configuration.

\begin{table}[H]
\centering
\caption{hyperparameter settings in \algo{} (\texttt{GCN}).}
\label{tab:hyperparams-GCN}
\vspace{-2ex}
\addtolength{\tabcolsep}{-0.3em}
\begin{small}
\resizebox{\columnwidth}{!}{%
\begin{tabular}{lcccccccc}
\toprule
Dataset   & $\beta$ & $\varepsilon$ & $T$ & \#layers & learning rate & hidden dim. & weight decay & dropout \\
\midrule
\textit{Cora}    &  0.003&   0.004&  3&   3&   0.001&  512& 5e-4& 0.8\\
\textit{CiteSeer}   &  0.008&   0.003&  3&   2&   0.001&  512& 1e-2& 0.5\\
\textit{PubMed}   &  0.008&   0.004&  4&   2&   0.001&  256& 5e-4& 0.7\\
\textit{Computer}   &  0.004&   0.004&  3&   3&   0.001&  512& 5e-5& 0.5\\
\textit{Photo}    &  0.007&   0.004&  3&   2&   0.001&  256& 5e-4& 0.6\\
\textit{CS}      &  0.003&   0.004&  3&   2&   0.001&  512& 5e-4& 0.3\\
\textit{Physics}   &  0.005&   0.004&  3&   2&   0.001&  64& 5e-4& 0.4\\
\bottomrule
\end{tabular}
}
\end{small}
\end{table}

\begin{table}[H]
\centering
\caption{hyperparameter settings in \algo{} (\texttt{GAT}).}
\label{tab:hyperparams-GAT}
\vspace{-2ex}
\addtolength{\tabcolsep}{-0.3em}
\begin{small}
\resizebox{\columnwidth}{!}{%
\begin{tabular}{lcccccccc}
\toprule
Dataset   & $\beta$ & $\varepsilon$ & $T$ & \#layers & learning rate & hidden dim. & weight decay & dropout \\
\midrule
\textit{Cora}    &  0.003&   0.04&  3&   3&   0.001&  256& 5e-4& 0.7\\
\textit{CiteSeer}   &  0.002&   0.04&  3&   3&   0.001&  256& 1e-2& 0.5\\
\textit{PubMed}   &  0.008&   0.04&  3&   2&   0.01&  512& 5e-4& 0.5\\
\textit{Computer}   &  0.005&   0.03&  1&   2&   0.001&  256& 5e-5& 0.7\\
\textit{Photo}    &  0.003&   0.04&  3&   3&   0.001&  64& 5e-5& 0.5\\
\textit{CS}      &  0.005&   0.04&  3&   2&   0.001&  256& 5e-4& 0.5\\
\textit{Physics}   &  0.004&   0.04&  3&   2&   0.001&  128& 5e-4& 0.6\\
\bottomrule
\end{tabular}
}
\end{small}
\end{table}

\begin{table}[H]
\centering
\caption{hyperparameter settings in \algo{} (\texttt{GraphSAGE}).}
\label{tab:hyperparams-SAGE}
\vspace{-2ex}
\addtolength{\tabcolsep}{-0.3em}
\begin{small}
\resizebox{\columnwidth}{!}{%
\begin{tabular}{lcccccccc}
\toprule
Dataset   & $\beta$ & $\varepsilon$ & $T$ & \#layers & learning rate & hidden dim. & weight decay & dropout \\
\midrule
\textit{Cora}    &  0.005&   0.04&  3&   3&   0.001&  256& 5e-4& 0.7
\\
\textit{CiteSeer}   &  0.005&   0.04&  3&   3&   0.001&  512& 1e-2& 0.7
\\
\textit{PubMed}   &  0.004&   0.04&  3&   4&   0.001&  512& 5e-4& 0.7
\\
\textit{Computer}   &  0.005&   0.04&  3&   4&   0.001&  128& 5e-5& 0.3
\\
\textit{Photo}    &  0.005&   0.04&  3&   6&   0.001&  64& 5e-5& 0.2
\\
\textit{CS}      &  0.001&   0.04&  3&   2&   0.001&  512& 5e-4& 0.5
\\
\textit{Physics}   &  0.003&   0.04&  3&   2&   0.001&  64& 5e-4& 0.7\\
\bottomrule
\end{tabular}
}
\end{small}
\end{table}

\begin{table}[H]
\centering
\caption{hyperparameter settings in \algo{} (\texttt{GCNII}).}
\label{tab:hyperparams-GCNII}
\vspace{-2ex}
\addtolength{\tabcolsep}{-0.3em}
\begin{small}
\resizebox{\columnwidth}{!}{%
\begin{tabular}{lcccccccc}
\toprule
Dataset   & $\beta$ & $\varepsilon$ & $T$ & \#layers & learning rate & hidden dim. & weight decay & dropout \\
\midrule
\textit{Cora}    &  0.003&   0.04&  3&   32&   0.001&  256& 5e-4& 0.6\\
\textit{CiteSeer}   &  0.003&   0.04&  3&   32&   0.001&  512& 5e-4& 0.7\\
\textit{PubMed}   &  0.004&   0.04&  3&   32&   0.005&  256& 5e-4& 0.6\\
\textit{Computer}   &  0.003&   0.04&  3&   2&   0.01&  256& 5e-5& 0.6\\
\textit{Photo}    &  0.008&   0.04&  3&   2&   0.01&  512& 5e-4& 0.6\\
\textit{CS}      &  0.002&   0.04&  3&   32&   0.005&  256& 5e-4& 0.6\\
\textit{Physics}   &  0.005&   0.04&  3&   32&   0.005&  256& 5e-4& 0.5\\
\bottomrule
\end{tabular}
}
\end{small}
\end{table}

\section{Baselines}
The code for all baseline methods was obtained from the repositories provided by the respective authors. Table~\ref{tab:links_code} summarizes the links to their implementations.

\begin{table}[H]
\centering
\caption{Links to code of baseline methods.}
\label{tab:links_code}
\vspace{-2ex}
\begin{small}
\begin{tabular}{l|p{0.7\columnwidth}}
\toprule
{\bf Method} & {\bf Link to code} \\
\midrule
GCN & \url{https://github.com/tkipf/gcn} \\
GCN* & \url{https://github.com/LUOyk1999/tunedGNN} \\
GAT & \url{https://github.com/PetarV-/GAT} \\
GAT* & \url{https://github.com/LUOyk1999/tunedGNN} \\
GraphSAGE & \url{https://github.com/williamleif/GraphSAGE} \\
GraphSAGE* & \url{https://github.com/LUOyk1999/tunedGNN} \\
SGC & \url{https://github.com/Tiiiger/SGC} \\
GCNII & \url{https://github.com/chennnM/GCNII} \\
APPNP & \url{https://github.com/benedekrozemberczki/APPNP} \\
DAGNN & \url{https://github.com/vthost/DAGNN} \\
JKNet & \url{https://github.com/ShinKyuY/Representation_Learning_on_Graphs_with_Jumping_Knowledge_Networks} \\
GPR-GNN & \url{https://github.com/jianhao2016/GPRGNN} \\
GRAND & \url{https://github.com/THUDM/GRAND} \\
GraphMix & \url{https://github.com/vikasverma1077/GraphMix} \\
GAM & \url{https://github.com/tensorflow/neural-structured-learning/tree/master/research/gam} \\
Violin & \url{https://github.com/XsLangley/Violin-IJCAI2023} \\
MGCN & \url{https://github.com/yiqun-wang/MGCN} \\
Meta-PN & \url{https://github.com/kaize0409/Meta-PN} \\
M3S & \url{https://github.com/datake/M3S} \\
DND-Net & \url{https://github.com/kaize0409/DND-NET} \\
NormProp & \url{https://github.com/Pallaksch/NormProp/tree/main} \\
PTA & \url{https://github.com/DongHande/PT_propagation_then_training} \\
\bottomrule
\end{tabular}
\end{small}
\end{table}

\input{figs/iteration}

\section{Additional Parameter Analyses}
Here, we investigate the influence of another key hyperparameter in SKN, namely the number of iterations $T$. 
To study the impact of $T$, we fix $\varepsilon = 0.04$ and vary $T$ in the range $\{1, 2, 3, 4, 5, 8, 10, 20, 30\}$.
Figures~\ref{fig:iterations-GCN} and ~\ref{fig:iterations-GAT} display the node classification performance obtained by \algo{} (\texttt{GCN}) and \algo{} (\texttt{GAT}) on all seven datasets, respectively.
For the figures, it can be observed that increasing the number of iterations $T$ initially leads to better performance as we successfully reduce the noise of over-confident pseudo labels, but encounters diminishing returns and an accuracy drop when $T$ becomes too large. Intuitively, when we perform numerous iterations of SKN, the pseudo-labels obtained will turn to be overly uniform, which will, in contrast, impair the model's performance. This suggests that while a moderate number of iterations, i.e., $T=3$ or $4$, is conducive for approximating a balanced pseudo-label distribution, excessive iterations may introduce noise or numerical instability.

%% file: figs/iteration.tex
\begin{figure*}[htbp]
\centering
\begin{small}
\resizebox{\textwidth}{!}{%
\subfloat[Cora]{
    \begin{tikzpicture}[scale=1]
    \begin{axis}[
        height=\columnwidth/2.5,
        width=\columnwidth/2.45,
        ylabel={\em Accuracy (\%)},
        xlabel={$T$},
        xmin=1, xmax=9,
        ymin=84.0, ymax=87,
        xtick={1,2,3,4,5,6,7,8,9},
        xticklabels={1,2,3,4,5,8,10,20,30},
        ytick={84.0,  85.0,  86.0, 87},
        yticklabel style = {font=\tiny},
        xticklabel style = {font=\tiny},
        every axis label/.style={font=\footnotesize},
        title style={font=\footnotesize},
        grid=major,
        grid style={dashed, gray!30}
    ]
    \addplot[line width=0.4mm, mark=o, color=purple]
        plot coordinates {
            (1, 84.90)
            (2, 84.84)
            (3, 86.16)
            (4, 84.46)
            (5, 85.00)
            (6, 84.52)
            (7, 85.38)
            (8, 84.86)
            (9, 84.58)
        };
    \end{axis}
    \end{tikzpicture}
}
\subfloat[CiteSeer]{
    \begin{tikzpicture}[scale=1]
    \begin{axis}[
        height=\columnwidth/2.5,
        width=\columnwidth/2.45,
        xlabel={$T$},
        xmin=1, xmax=9,
        ymin=73, ymax=76,
        xtick={1,2,3,4,5,6,7,8,9},
        xticklabels={1,2,3,4,5,8,10,20,30},
        yticklabel style = {font=\tiny},
        xticklabel style = {font=\tiny},
        every axis label/.style={font=\footnotesize},
        title style={font=\footnotesize},
        grid=major,
        grid style={dashed, gray!30}
    ]
    \addplot[line width=0.4mm, mark=o, color=purple]
        coordinates {
            (1, 75.40)
            (2, 75.50)
            (3, 75.53)
            (4, 75.10)
            (5, 74.83)
            (6, 74.23)
            (7, 73.33)
            (8, 73.27)
            (9, 73.33)
        };
    \end{axis}
    \end{tikzpicture}
}
\subfloat[PubMed]{
    \begin{tikzpicture}[scale=1]
    \begin{axis}[
        height=\columnwidth/2.5,
        width=\columnwidth/2.45,
        xlabel={$T$},
        xmin=1, xmax=9,
        ymin=79, ymax=82,
        xtick={1,2,3,4,5,6,7,8,9},
        xticklabels={1,2,3,4,5,8,10,20,30},
        yticklabel style = {font=\tiny},
        xticklabel style = {font=\tiny},
        every axis label/.style={font=\footnotesize},
        title style={font=\footnotesize},
        grid=major,
        grid style={dashed, gray!30}
    ]
    \addplot[line width=0.4mm, mark=o, color=purple]
        coordinates {
            (1, 81.20)
            (2, 81.07)
            (3, 81.22)
            (4, 81.27)
            (5, 81.07)
            (6, 80.77)
            (7, 80.70)
            (8, 80.83)
            (9, 79.40)
        };
    \end{axis}
    \end{tikzpicture}
}
\subfloat[Computer]{
    \begin{tikzpicture}[scale=1]
    \begin{axis}[
        height=\columnwidth/2.5,
        width=\columnwidth/2.45,
        xlabel={ $T$},
        xmin=1, xmax=9,
        ymin=80, ymax=83,
        xtick={1,2,3,4,5,6,7,8,9},
        xticklabels={1,2,3,4,5,8,10,20,30},
        yticklabel style = {font=\tiny},
        xticklabel style = {font=\tiny},
        every axis label/.style={font=\footnotesize},
        title style={font=\footnotesize},
        grid=major,
        grid style={dashed, gray!30}
    ]
    \addplot[line width=0.4mm, mark=o, color=purple]
        coordinates {
            (1, 82.28)
            (2, 81.91)
            (3, 82.79)
            (4, 81.85)
            (5, 80.47)
            (6, 81.51)
            (7, 81.31)
            (8, 82.21)
            (9, 82.07)
        };
    \end{axis}
    \end{tikzpicture}
}
\subfloat[Photo]{
    \begin{tikzpicture}[scale=1]
    \begin{axis}[
        height=\columnwidth/2.5,
        width=\columnwidth/2.45,
        xlabel={$T$},
        xmin=1, xmax=9,
        ymin=89, ymax=92,
        xtick={1,2,3,4,5,6,7,8,9},
        xticklabels={1,2,3,4,5,8,10,20,30},
        ytick={89,  90,  91,  92},
        yticklabel style = {font=\tiny},
        xticklabel style = {font=\tiny},
        every axis label/.style={font=\footnotesize},
        title style={font=\footnotesize},
        grid=major,
        grid style={dashed, gray!30},
        ylabel={}
    ]
    \addplot[line width=0.4mm, mark=o, color=purple]
        plot coordinates {
            (1, 89.46)
            (2, 90.03)
            (3, 91.39)
            (4, 90.12)
            (5, 89.97)
            (6, 90.45)
            (7, 90.16)
            (8, 90.55)
            (9, 90.03)
        };
    \end{axis}
    \end{tikzpicture}
}
\subfloat[CS]{
    \begin{tikzpicture}[scale=1]
    \begin{axis}[
        height=\columnwidth/2.5,
        width=\columnwidth/2.45,
        xlabel={$T$},
        xmin=0.0, xmax=10.0,
        ymin=85, ymax=95,
        xtick={1,2,3,4,5,6,7,8,9},
        xticklabels={1,2,3,4,5,8,10,20,30},
        xticklabel style = {font=\tiny},
        every axis label/.style={font=\footnotesize},
        title style={font=\footnotesize},
        grid=major,
        grid style={dashed, gray!30}
    ]
    \addplot[line width=0.4mm, mark=o, color=purple]
        coordinates {
            (1, 91.09)
            (2, 92.05)
            (3, 93.68)
            (4, 92.85)
            (5, 92.97)
            (6, 92.79)
            (7, 92.77)
            (8, 90.40)
            (9, 88.24)
        };
    \end{axis}
    \end{tikzpicture}
}
\subfloat[Physics]{
    \begin{tikzpicture}[scale=1]
    \begin{axis}[
        height=\columnwidth/2.5,
        width=\columnwidth/2.45,
        xlabel={$T$},
        xmin=1, xmax=9,
        ymin=80, ymax=95,
        xtick={1,2,3,4,5,6,7,8,9},
        xticklabels={1,2,3,4,5,8,10,20,30},
        yticklabel style = {font=\tiny},
        xticklabel style = {font=\tiny},
        every axis label/.style={font=\footnotesize},
        title style={font=\footnotesize},
        grid=major,
        grid style={dashed, gray!30}
    ]
    \addplot[line width=0.4mm, mark=o, color=purple]
        coordinates {
            (1, 83.61)
            (2, 94.21)
            (3, 94.40)
            (4, 94.22)
            (5, 94.20)
            (6, 93.47)
            (7, 93.24)
            (8, 83.14)
            (9, 80.15)
        };
    \end{axis}
    \end{tikzpicture}
}
}
\end{small}
\vspace{-2ex}
\caption{Varying $T$ in \algo{} (\texttt{GCN}).}
\label{fig:iterations-GCN}
\vspace{-2ex}
\end{figure*}

\begin{figure*}[htbp]\centering
\begin{small}
\resizebox{\textwidth}{!}{%
\subfloat[Cora]{
    \begin{tikzpicture}[scale=1]
    \begin{axis}[
        height=\columnwidth/2.5,
        width=\columnwidth/2.45,
        ylabel={\em Accuracy (\%)},
        xlabel={$T$},
        xmin=1, xmax=9,
        ymin=82.5, ymax=85,
        xtick={1,2,3,4,5,6,7,8,9},
        xticklabels={1,2,3,4,5,8,10,20,30},
        ytick={82.5, 83, 83.5, 84, 84.5, 85},
        yticklabel style = {font=\tiny},
        xticklabel style = {font=\tiny},
        every axis label/.style={font=\footnotesize},
        title style={font=\footnotesize},
        grid=major,
        grid style={dashed, gray!30}
    ]
    \addplot[line width=0.4mm, mark=o, color=purple]
        coordinates {
            (1, 84.62)
            (2, 84.44)
            (3, 84.72)
            (4, 83.72)
            (5, 83.78)
            (6, 83.44)
            (7, 83.54)
            (8, 82.88)
            (9, 82.82)
        };
    \end{axis}
    \end{tikzpicture}
}
\subfloat[CiteSeer]{
    \begin{tikzpicture}[scale=1]
    \begin{axis}[
        height=\columnwidth/2.5,
        width=\columnwidth/2.45,
        xlabel={$T$},
        xmin=1, xmax=9,
        ymin=73, ymax=75,
        xtick={1,2,3,4,5,6,7,8,9},
        xticklabels={1,2,3,4,5,8,10,20,30},
        yticklabel style = {font=\tiny},
        xticklabel style = {font=\tiny},
        every axis label/.style={font=\footnotesize},
        title style={font=\footnotesize},
        grid=major,
        grid style={dashed, gray!30}
    ]
    \addplot[line width=0.4mm, mark=o, color=purple]
        coordinates {
            (1, 74.18)
            (2, 74.30)
            (3, 74.46)
            (4, 74.30)
            (5, 74.20)
            (6, 74.34)
            (7, 74.32)
            (8, 74.50)
            (9, 74.28)
        };
    \end{axis}
    \end{tikzpicture}
}
\subfloat[PubMed]{
    \begin{tikzpicture}[scale=1]
    \begin{axis}[
        height=\columnwidth/2.5,
        width=\columnwidth/2.45,
        xlabel={$T$},
        xmin=1, xmax=9,
        ymin=79, ymax=80,
        xtick={1,2,3,4,5,6,7,8,9},
        xticklabels={1,2,3,4,5,8,10,20,30},
        yticklabel style = {font=\tiny},
        xticklabel style = {font=\tiny},
        every axis label/.style={font=\footnotesize},
        title style={font=\footnotesize},
        grid=major,
        grid style={dashed, gray!30}
    ]
    \addplot[line width=0.4mm, mark=o, color=purple]
        coordinates {
            (1, 79.34)
            (2, 79.36)
            (3, 79.50)
            (4, 79.34)
            (5, 79.38)
            (6, 79.36)
            (7, 79.32)
            (8, 79.32)
            (9, 79.32)
        };
    \end{axis}
    \end{tikzpicture}
}
\subfloat[Computer]{
    \begin{tikzpicture}[scale=1]
    \begin{axis}[
        height=\columnwidth/2.5,
        width=\columnwidth/2.45,
        xlabel={$T$},
        xmin=1, xmax=9,
        ymin=77, ymax=85,
        xtick={1,2,3,4,5,6,7,8,9},
        xticklabels={1,2,3,4,5,8,10,20,30},
        yticklabel style = {font=\tiny},
        xticklabel style = {font=\tiny},
        every axis label/.style={font=\footnotesize},
        title style={font=\footnotesize},
        grid=major,
        grid style={dashed, gray!30}
    ]
    \addplot[line width=0.4mm, mark=o, color=purple]
        coordinates {
            (1, 83.52)
            (2, 83.03)
            (3, 83.33)
            (4, 82.39)
            (5, 81.77)
            (6, 80.60)
            (7, 81.20)
            (8, 78.24)
            (9, 79.13)
        };
    \end{axis}
    \end{tikzpicture}
}
\subfloat[Photo]{
    \begin{tikzpicture}[scale=1]
    \begin{axis}[
        height=\columnwidth/2.5,
        width=\columnwidth/2.45,
        xlabel={$T$},
        xmin=1, xmax=9,
        ymin=88, ymax=93,
        xtick={1,2,3,4,5,6,7,8,9},
        xticklabels={1,2,3,4,5,8,10,20,30},
        yticklabel style = {font=\tiny},
        xticklabel style = {font=\tiny},
        every axis label/.style={font=\footnotesize},
        title style={font=\footnotesize},
        grid=major,
        grid style={dashed, gray!30},
        ylabel={}
    ]
    \addplot[line width=0.4mm, mark=o, color=purple]
        coordinates {
            (1, 92.48)
            (2, 92.29)
            (3, 92.54)
            (4, 91.72)
            (5, 91.52)
            (6, 90.16)
            (7, 88.65)
            (8, 89.08)
            (9, 88.51)
        };
    \end{axis}
    \end{tikzpicture}
}
\subfloat[CS]{
    \begin{tikzpicture}[scale=1]
    \begin{axis}[
        height=\columnwidth/2.5,
        width=\columnwidth/2.45,
        xlabel={$T$},
        xmin=1, xmax=9,
        ymin=91.40, ymax=93.20,
        xtick={1,2,3,4,5,6,7,8,9},
        xticklabels={1,2,3,4,5,8,10,20,30},
        xticklabel style = {font=\tiny},
        every axis label/.style={font=\footnotesize},
        title style={font=\footnotesize},
        grid=major,
        grid style={dashed, gray!30}
    ]
    \addplot[line width=0.4mm, mark=o, color=purple]
        coordinates {
            (1, 92.86)
            (2, 92.97)
            (3, 93.07)
            (4, 92.84)
            (5, 92.70)
            (6, 92.55)
            (7, 92.17)
            (8, 91.83)
            (9, 91.49)
        };
    \end{axis}
    \end{tikzpicture}
}
\subfloat[Physics]{
    \begin{tikzpicture}[scale=1]
    \begin{axis}[
        height=\columnwidth/2.5,
        width=\columnwidth/2.45,
        xlabel={$T$},
        xmin=1, xmax=9,
        ymin=93.75, ymax=93.90,
        xtick={1,2,3,4,5,6,7,8,9},
        xticklabels={1,2,3,4,5,8,10,20,30},
        yticklabel style = {font=\tiny},
        xticklabel style = {font=\tiny},
        every axis label/.style={font=\footnotesize},
        title style={font=\footnotesize},
        grid=major,
        grid style={dashed, gray!30}
    ]
    \addplot[line width=0.4mm, mark=o, color=purple]
        coordinates {
            (1, 93.88)
            (2, 93.84)
            (3, 93.88)
            (4, 93.88)
            (5, 93.80)
            (6, 93.88)
            (7, 93.83)
            (8, 93.85)
            (9, 93.85)
        };
    \end{axis}
    \end{tikzpicture}
}
}
\end{small}
\vspace{-2ex}
\caption{Varying $T$ in \algo{} (\texttt{GAT}).}
\label{fig:iterations-GAT}
\vspace{0ex}
\end{figure*}